\documentclass[fleqn,10pt]{wlpeerj}
\usepackage{amssymb}
\usepackage{amsmath,graphicx,amssymb,bm,textcomp}
\usepackage[%
unicode,naturalnames,
breaklinks=true,
colorlinks=true,
linkcolor=black,
anchorcolor=black,
citecolor=black,
filecolor=black,
menucolor=black,
runcolor=black,
linktocpage,
draft=false,
pageanchor=true,
hyperfigures=true,
plainpages=false
]{hyperref}
\usepackage{color}
\usepackage{multirow,multicol}
\usepackage{tikz,rotating}
\usepackage{pifont}
\usepackage{caption}
\usepackage{stfloats}
\usepackage{tabularx,ragged2e}
\usepackage{balance}
\usepackage{tcolorbox}
\usepackage{subcaption}
\usepackage{array,etoolbox}

\usepackage[T1]{fontenc}
 
\usetikzlibrary{fit,calc}

\usepackage{xcolor}

\preto\tabular{\setcounter{magicrownumbers}{0}}
\newcounter{magicrownumbers}
\def\rownumber{}

\definecolor{usc}{rgb}{0.6,0.106,0.117}

\newcommand{\Pg}[1]{\noindent{{\color{red}\textbf{#1}}}}
\renewcommand{\Pg}[1]{}
\newcommand{\heading}[1]{\noindent{{\color{black}\textbf{#1:}}}}

\newcounter{hypcount}

\makeatletter
\newcommand\newtag[2]{#1\def\@currentlabel{#1}\label{#2}}
\makeatother

\title{Confusion2Vec: Towards enriching vector space word representations with representational ambiguities}

\author{Prashanth Gurunath Shivakumar}
\author{Panayiotis Georgiou}
\affil{Electrical and Computer Engineering, \\
  University of Southern California, Los Angeles, CA, United States}
\corrauthor{Panayiotis Georgiou}{georgiou@sipi.usc.edu}

\begin{abstract}
Word vector representations are a crucial part of Natural Language Processing and Human Computer Interaction.
In this paper, we propose a novel word vector representation, Confusion2Vec, motivated from the human speech production and perception that encodes representational ambiguity. 
Humans employ both acoustic similarity cues and contextual cues to decode information and we focus on a model that incorporates both sources of information.
The representational ambiguity of acoustics, which manifests itself in word confusions, is often resolved by both humans and machines through contextual cues.
A range of representational ambiguities can emerge in various domains further to acoustic perception, such as morphological transformations, word segmentation, paraphrasing for natural language processing tasks like machine translation etc.
In this work, we present a case study in application to Automatic Speech Recognition (ASR) task, where the word representational ambiguities/confusions are related to acoustic similarity.
We present several techniques to train an acoustic perceptual similarity representation ambiguity.
We term this Confusion2Vec and learn on unsupervised-generated data from Automatic Speech Recognition confusion networks or lattice-like structures.
Appropriate evaluations for the Confusion2Vec are formulated for gauging acoustic similarity in addition to semantic-syntactic and word similarity evaluations.
The Confusion2Vec is able to model word confusions efficiently, without compromising on the semantic-syntactic word relations, thus effectively enriching the word vector space with extra task relevant ambiguity information.
We provide an intuitive exploration of the 2-dimensional Confusion2Vec space using Principal Component Analysis of the embedding and relate to semantic relationships, syntactic relationships and acoustic relationships.
We show through this that the new space preserves the semantic/syntactic relationships while robustly encoding acoustic similarities.
The potential of the new vector representation and its ability in the utilization of uncertainty information associated with the lattice is demonstrated through small examples relating to the task of ASR error correction.
\end{abstract}

\begin{document}
\flushbottom

\maketitle
\thispagestyle{empty}

\section{Introduction}\label{sec:intro}
\Pg{Need for Word vector representation}
Decoding human language is challenging for machines.
It involves estimation of efficient, meaningful representation of words.
Machines represent the words in the form of real vectors and the language as a vector space.
Vector space representations of language have applications spanning natural language processing (NLP) and human computer interaction (HCI) fields.
More specifically, word embeddings can act as features for Machine Translation, Automatic Speech Recognition, Document Topic Classification, Information Retrieval, Sentiment Classification, Emotion Recognition, Behavior Recognition, Question Answering etc.

\Pg{Earlier Feature Representation}
Early work employed words as the fundamental unit of feature representation.
This could be thought of as each word representing an orthogonal vector in a n-dimensional vector space of language with n-words (often referred to as one-hot representation).
Such a representation, due to the inherent orthogonality, lacks crucial information regarding inter-word relationships such as similarity.
Several techniques found using co-occurrence information of words to be a better feature representation (Ex: n-gram Language Modeling).

\Pg{Matrix Factorization Methods}
Subsequent studies introduced few matrix factorization based techniques to estimate a more efficient, reduced dimensional vector space based on word co-occurrence information.
Latent Semantic Analysis (LSA) assumes an underlying vector space spanned by orthogonal set of latent variables closely associated with the semantics/meanings of the particular language.
The dimension of this vector space is much smaller than the one-hot representation \citep{deerwester1990indexing}.
LSA was proposed initially for information retrieval and indexing, but soon gained popularity for other NLP tasks.
\cite{hofmann1999probabilistic} proposed Probabilistic LSA replacing the co-occurrence information by a statistical class based model leading to better vector space representations. 

Another popular matrix factorization method, the Latent Dirichlet Allocation (LDA) assumes a generative statistical model where the documents are characterized as a mixture of latent variables representing topics which are described by word distributions \citep{blei2003latent}.

\Pg{Neural Network based Techniques}
Recently neural networks gained popularity. They often outperform the N-gram models \citep{bengio2003neural,mikolov2010recurrent} and enable estimation of more complex models incorporating much larger data than before.
Various neural network based vector space estimation of words were proposed. 
\cite{bengio2003neural} proposed feed-forward neural network based language models which jointly learned the distributed word representation along with the probability distribution associated with the representation.
Estimating a reduced dimension continuous word representation allows for efficient probability modeling, thereby resulting in much lower perplexity compared to an n-gram model.
Recurrent neural network based language models, with inherent memory, allowed for the exploitation of much longer context, providing further improvements compared to feed forward neural networks \citep{mikolov2010recurrent}.

\Pg{Word2Vec}
\cite{mikolov2013efficient} proposes a new technique of estimating vector representation (popularly termed word2vec) which showed promising results in preserving the semantic and syntactic relationships between words.
Two novel architectures based on simple log-linear modeling (i) continuous skip-gram and (ii) continuous bag-of-words are introduced.
Both the models are trained to model local context of word occurrences.
The continuous skip-gram model predicts surrounding words given the current word. Whereas, the continuous bag-of-words model predicts the current word given its context.
The task evaluation is based on answering various analogy questions testing semantic and syntactic word relationships.
Several training optimizations and tips were proposed to further improve estimation of the vector space by \cite{mikolov2013distributed, mnih2013learning}.
Such efficient representation of words directly influences the performance of NLP tasks like sentiment classification \citep{kim2014convolutional}, part-of-speech tagging \citep{ling2015two}, text classification \citep{lilleberg2015support,joulin2016bag}, document categorization \citep{xing2014document} and many more.

\Pg{Word2Vec Extensions}
Subsequent research efforts on extending word2vec involve expanding the word representation to phrases \citep{mikolov2013distributed}, sentences and documents \citep{le2014distributed}. 
Similarly, training for contexts derived from syntactic dependencies of a word is shown to produce useful representations \citep{levy2014dependency}.
Using morphemes for word representations can enrich the vector space and provide gains especially for unknown, rarely occurring, complex words and morphologically rich languages \citep{luong2013better,botha2014compositional,qiu2014co,cotterell2015morphological,soricut2015unsupervised}.
Likewise, incorporating sub-word representations of words for the estimation of vector space is beneficial \citep{bojanowski2017enriching}.
Similar studies using characters of words have also been tried \citep{chen2015joint}.
\cite{YinS16} explored ensemble techniques for exploiting complementary information over multiple word vector spaces.
Studies by \cite{mikolov2013exploiting,faruqui2014improving} demonstrate that vector space representations are extremely useful in extending the model from one language to another (or multi-lingual extensions) since the semantic relations between words are invariant across languages.

\Pg{Other relevant works}
Some have tried to combine the advantages from both matrix factorization based techniques and local-context word2vec models.
\cite{pennington2014glove} proposes global log-bilinear model for modeling global statistical information as in the case of global matrix factorization techniques along with the local context information as in the case of word2vec.

\Pg{Proposed Work}
The goal of this study is to come up with a new vector space representation for words which incorporates the uncertainty information in the form of word confusions present in lattice like structures (e.g. confusion networks).
Here, the word confusions refers to any word level ambiguities resultant of perception confusability or any algorithms such as machine translation, ASR etc.
For example, acoustically confusable words in ASR lattices: "two" and "to" (see Figure~\ref{fig:asr_sausage}).
A word lattice is a compact representation (directed acyclic weighted graphs) of different word sequences that are likely possible.
A confusion network is a special type of lattice, where each word sequence is made to pass through each node of the graph.
The lattices and confusion networks embed word confusion information.
The study takes motivation from human perception, i.e., the ability of humans to decode information based on two fairly independent information streams (see Section~\ref{sec:human} for examples): (i) linguistic context (modeled by word2vec like word vector representations), and (ii) acoustic confusability (relating to phonology).

The present word vector representations like word2vec only incorporate the contextual confusability during modeling.
Hence, in order to handle confusability and to decode human language/speech successfully, there is a need to model both the dimensions.
Although, primarily, the motivation is derived from human speech and perception, the confusions are not constrained to acoustics and can be extended to any confusions parallel to the linguistic contexts, for example, confusions present in lattices.
Most of the machine learning algorithms output predictions as a probability measure. 
This uncertainty information stream can be expressed in the form of a lattice or a confusion network temporally, and is often found to contain useful information for subsequent processing and analysis.
The scope of this work is to introduce a complementary (ideally orthogonal) subspace in addition to the underlying word vector space representation captured by word2vec.
This new subspace captures the word confusions orthogonal to the syntactic and semantics of the language.
We propose Confusion2Vec vector space operating on lattice like structures, specifically word confusion networks.
We introduce several training configurations and evaluate their effectiveness. 
We also formulate appropriate evaluation criterion to assess the performance of each orthogonal subspaces, first independently and then jointly.
Analysis of the proposed word vector space representation is carried out.

\Pg{Paper Structure}
The rest of the paper is organized as follows. Motivation for Confusion2vec, i.e., the need to model word-confusions for word embeddings, is provided through means of human speech \& perception, machine learning, and through potential applications in section~\ref{sec:motivation}. 
A particular case study is chosen and the problem is formulated in section~\ref{sec:case_study}. 
In section~\ref{sec:proposed}, different training configurations for efficient estimation of word embeddings are proposed. 
Additional tuning schemes for the proposed Confusion2vec models are presented in section~\ref{sec:tuning}.
Evaluation criterion formulation and evaluation database creation is presented in section~\ref{sec:evaluation}. 
Experimental setup and baseline system is described in section~\ref{sec:baseline_setup}. 
Results are tabulated and discussed in section~\ref{sec:results}. Word vector space analysis is performed and findings are presented in section~\ref{sec:analysis}. 
Section~\ref{sec:discussion} discusses with the help of few toy examples, the benefits of the Confusion2vec embeddings for the task of ASR error correction.
Section~\ref{sec:conclusion} draws the conclusion of the study and finally the future research directions are discussed in Section~\ref{sec:future_work}.

\section{Motivation}\label{sec:motivation}

One efficient way to represent words as vectors is to represent them in a space that preserves the semantic and syntactic relations between the words in the language.
Word2vec describes a technique to achieve such a representation by trying to predict the current word from its local context (or vice-versa) over a large text corpora.
The estimated word vectors are shown to encode efficient syntactic-semantic language information.
In this work we propose a new vector space for word representation which incorporates various forms of word confusion information in addition to the semantic \& syntactic information.
The new vector space is inspired and motivated from the following factors from human speech production \& perception and machine learning.

\subsection{Human speech production, perception and hearing}\label{sec:human}
In our every day interactions, confusability can often result in the need for context to decode the underlying words.
\begin{equation}
``\text{Please }\rule{0.75cm}{0.15mm}\text{ a seat.}'' \tag{Example 1}\label{ex:1}
\end{equation}
In~\ref{ex:1}, the missing word could be guessed from its context and narrowed down to either ``have'' or ``take''.
This context information is modeled through language models.
More complex models such as word2vec also use the contextual information to model word vector representations.

On the other hand, confusability can also originate from other sources such as acoustic representations.
\begin{equation}
``\text{I want to \underline{seat}}'' \tag{Example 2}\label{ex:2}
\end{equation}
In~\ref{ex:2}, the underlined word is mispronounced/misheard, and grammatically incorrect.
In this case, considering the context there exists a lot of possible correct substitutions for the word ``seat'' and hence the context is less useful.
The acoustic construct of the word ``seat'' can present additional information in terms of acoustic alternatives/similarity, such as ``sit'' and ``seed''.
\begin{equation}
``\text{I want to \underline{s---}}'' \tag{Example 3}\label{ex:3}
\end{equation}
Similarly in~\ref{ex:3}, the underlined word is incomplete. 
The acoustic confusability information can be useful in the above case of broken words.
Thus, since the confusability is acoustic, purely lexical vector representations like word2vec fail to encode or capture it.
In this work, we propose to additionally encode the word (acoustic) confusability information to learn a better word embedding.
Although the motivation is specific to acoustics in this case, it could be extended to other inherent sources of word-confusions spanning various machine learning applications.

\subsection{Machine Learning Algorithms}\label{sec:machine}
Most of the machine learning algorithms output hypothesis as a probability measure.
Such a hypothesis could be represented in the form of a lattice, confusion network or n-best lists.
It is often useful to consider the uncertainty associated with the hypothesis for subsequent processing and analysis (see Section~\ref{sec:apps} for potential applications).
The uncertainty information is often, orthogonal to the contextual dimension and is specific to the task attempted by the machine learning algorithms.

Along this direction, recently, there have been several efforts concentrated on introducing lattice information into the neural network architecture. 
Initially, Tree-LSTM was proposed enabling tree-structured network topologies to be inputted to the RNNs \citep{tai2015improved}, which could be adapted and applied to lattices \citep{sperber2017neural}. 
LatticeRNN was proposed for processing word level lattices for ASR \citep{ladhak2016latticernn}. 
Lattice based Gated Recurrent Units (GRUs) \citep{su2017lattice} and lattice-to-sequence models \citep{tan2018lattice} were proposed for reading word lattice as input, specifically a lattice with tokenization alternatives for machine translation models. 
LatticeLSTM was adopted for lattice-to-sequence model incorporating lattice scores for the task of speech translation by \cite{sperber2017neural}. 
\cite{buckman2018neural} proposed Neural lattice language models which enables to incorporate many possible meanings for words and phrases (paraphrase alternatives).

Thus, a vector space representation capable of embedding relevant uncertainty information in the form of word confusions present in lattice-like structures or confusion networks along with the Semantic \& Syntactic can be potentially superior to word2vec space.

\section{Case Study: Application to Automatic Speech Recognition}\label{sec:case_study}
\begin{figure}[t]
\centering
\includegraphics[width=0.8\linewidth]{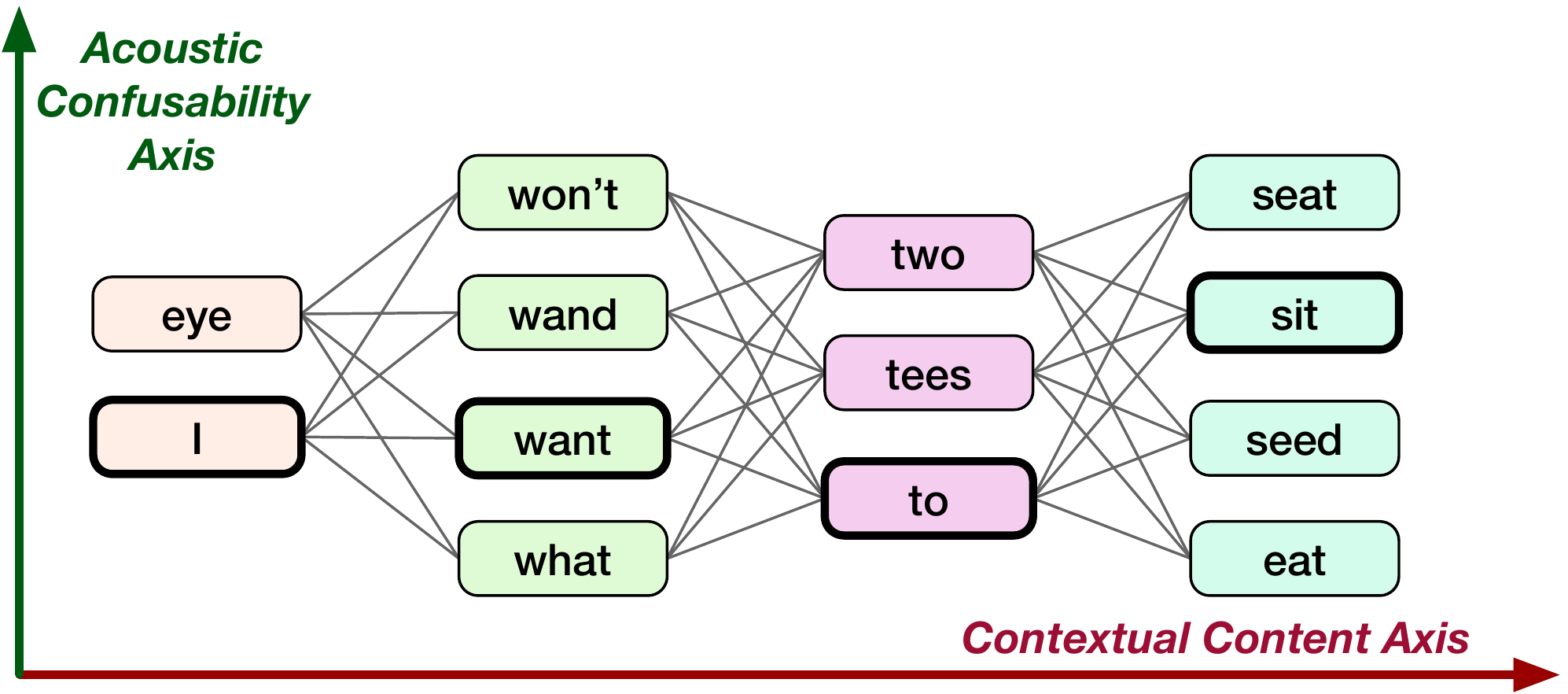}
\captionsetup{justification=centering}
\caption{\textbf{An example confusion network for ground-truth utterance ``I want to sit.''}}\label{fig:asr_sausage}
\end{figure}

In this work, we consider the ASR task as a case study to demonstrate the effectiveness of the proposed Confusion2vec model in modeling acoustic word-confusability. 
However, the technique can be adopted for a lattice or confusion network output from potentially any algorithm to capture various patterns as discussed in section~\ref{sec:apps}, in which case the confusion-subspace (vertical ambiguity in figure~\ref{fig:asr_sausage}), is no longer constrained to acoustic word-confusions.

An ASR lattice contains multiple paths over acoustically similar words. 
A lattice could be transformed and represented as a linear graph forcing every path to pass through all the nodes \citep{xue2005improved,mangu2000finding}. 
Such a linear graph is referred to as a confusion network. 
Figure~\ref{fig:asr_sausage} shows a sample confusion network output by ASR for the ground truth ``I want to sit''.
The confusion network could be viewed along two fundamental dimensions of information (see figure~\ref{fig:asr_sausage}): (i) Contextual axis - sequential structure of a sentence, (ii) Acoustic axis - similarly sounding word alternatives.
Traditional word vector representations such as word2vec only model the contextual information (the horizontal (red) direction in Figure~\ref{fig:asr_sausage}).
The word confusions, for example, the acoustic contextualization as in Figure~\ref{fig:asr_sausage} (the vertical (green) direction in Figure~\ref{fig:asr_sausage}) is not encoded.
We propose to additionally capture the co-occurrence information along the acoustic axis orthogonal to the word2vec.
This is the main focus of our work, i.e., to jointly learn the vertical, word-confusion context and the horizontal, semantic and syntactic context.
In other words, we hypothesize to derive relationships between the semantics and syntaxes of language and the word-confusions (acoustic-confusion).

\subsection{Related Work}
\cite{bengio2014} trained a continuous word embedding of acoustically alike words (using n-gram feature representation of words) to replace the state space models (HMMs), decision trees and lexicons of an ASR. Through the use of such an embedding and lattice re-scoring technique demonstrated improvements in word error rates of ASR.
The embeddings are also shown to be useful in application to the task of ASR error detection by \cite{Ghannay_2016}. 
A few evaluation strategies are also devised to evaluate phonetic and orthographic similarity of words.
Additionally, there have been studies concentrating on estimating word embeddings from acoustics \citep{kamper2016deep,chung2016audio,levin2013fixed,he2016multi} with evaluations based on acoustic similarity measures.
Parallely, word2vec like word embeddings have been used successfully to improve ASR Error detection performance \citep{7362668,Ghannay:2015:CCW:2963447.2963456}.
We believe the proposed exploitation of both information sources, i.e., acoustic relations and linguistic relations (semantics and syntaxes) will be beneficial in ASR and error detection, correction tasks.
The proposed confusion2vec operates on the lattice output of the ASR in contrast to the work on acoustic word embeddings \citep{kamper2016deep,chung2016audio,levin2013fixed,he2016multi} which is directly trained on audio.
The proposed Confusion2vec differs from works by \cite{bengio2014} and \cite{Ghannay_2016}, which also utilize audio data with the hypothesis that the layer right below softmax layer of a deep end-to-end ASR contains acoustic similarity information of words.
Confusion2vec can also be potentially trained without an ASR, on artificially generated data, emulating an ASR \citep{tan2010automatic,sagae2012hallucinated,celebi2012semi,kurata2011training,dikici2012performance,xu2012phrasal}.
Thus, Confusion2vec can potentially be trained in a completely unsupervised manner and with appropriate model parameterization incorporate various degrees of acoustic confusability, e.g. stemming from noise or speaker conditions. 

Further, in contrast to the prior works on lattice encoding RNNs \citep{tai2015improved,sperber2017neural,ladhak2016latticernn,su2017lattice,tan2018lattice,buckman2018neural}, which concentrate on incorporating the uncertainty information embedded in the word lattices by modifying the input architecture for recurrent neural network, we propose to introduce the ambiguity information from the lattices to the word embedding explicitly.
We expect similar advantages as with lattice encoding RNNs in using the pre-trained confusion2vec embedding towards various tasks like ASR, Machine translation etc.
Moreover, our architecture doesn't require memory which has significant advantages in terms of training complexity.
We propose to train the embedding in a similar way to word2vec models \citep{mikolov2013efficient}. 
All the well studied previous efforts towards optimization of training such models \citep{mikolov2013distributed, mnih2013learning}, should apply to our proposed model.

\section{Proposed Models}\label{sec:proposed}
In this section, we propose four training schemes for Confusion2Vec.
The training schemes are based on the word2vec model.
Word2vec work \citep{mikolov2013efficient} proposed log-linear models, i.e., neural network consisting of a single linear layer (projection matrix) without non-linearity. 
These models have significant advantages in training complexity. 
\cite{mikolov2013efficient} found the skip-gram model to be superior to the bag-of-word model in a semantic-syntactic analogy task.
Hence, we only employ the skip-gram configuration in this work.
Appropriately, the skip-gram word2vec model is also adopted as the baseline for this work.
The choice of the skip-gram modeling in this work is mainly based on its popularity, ease of implementation, low complexity and being a well-proven technique in the community.
However, we strongly believe the proposed concept (introducing word ambiguity information) is independent of the modeling technique itself and should translate to relatively newer techniques like GloVe \cite{pennington2014glove} and fastText \cite{bojanowski2017enriching}.

\subsection{Top-Confusion Training - C2V-1}

\begin{figure*}[t]
\centering
  \includegraphics[width=\linewidth]{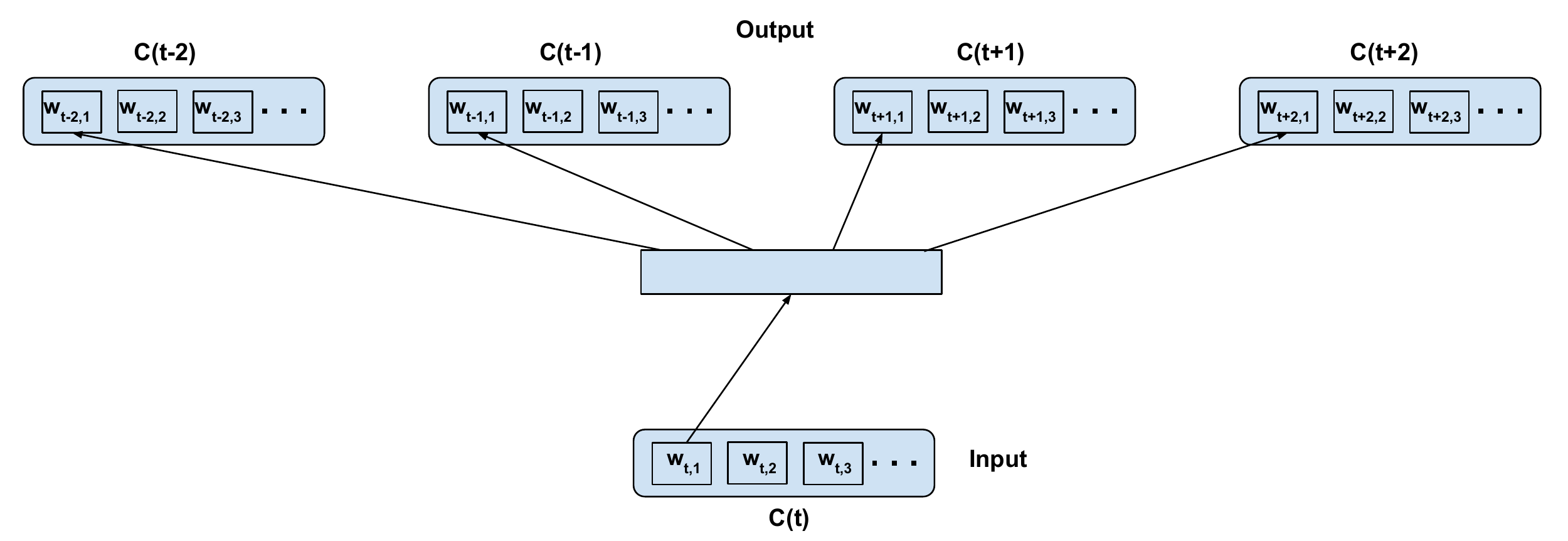}
	\captionsetup{justification=centering}
	\caption{\textbf{Top-Confusion2Vec Training scheme for Confusion networks.}\\
	\small{$c(t)$ is a unit word confusion in the confusion network at a time-stamp $t$, i.e., $c(t)$ represents a set of arcs between two adjacent nodes of a confusion network, representing a set of confusable words.\\
$w_{t,i}$ is the $i^{th}$ most probable word in the confusion $c(t)$.\\
Word confusions are sorted in decreasing order of their posterior probability: $P(w_{t,1})>P(w_{t,2})>P(w_{t,3})...$}}\label{fig:baseline_w2v}
\end{figure*}

We adapt the word2vec contextual modeling to operate on the confusion network (in our case confusion network of an ASR).
Figure~\ref{fig:baseline_w2v} shows the training configuration of the skip-gram word2vec model on the confusion network.
The top-confusion model considers the context of only the top hypothesis of the confusion network (single path) for training.
For clarity we call this the C2V-1 model since it's using only the 1 top hypothesis.
The words $w_{t-2,1}$, $w_{t-1,1}$, $w_{t+1,1}$ and $w_{t+2,1}$ (i.e., the most probable words in the confusions $C(t-2)$, $C(t-1)$, $C(t+1)$ and $C(t+2)$ respectively) are predicted from $w_{t,1}$ (i.e., the most probable word in $C(t)$) for a skip-window of 2 as depicted in Figure~\ref{fig:baseline_w2v}.
The top hypothesis typically consists of noisy transformations of the reference ground-truth (Note: the confusion network will inherently introduce additional paths to the lattice).
In the case of a confusion network of an ASR, the noisy transformations correspond to acoustic word confusions.
Thus, the top-confusion model implicitly captures word confusions (co-occurring within the context of the skip-window).

\begin{figure*}[t]
\centering
  \includegraphics[width=\linewidth]{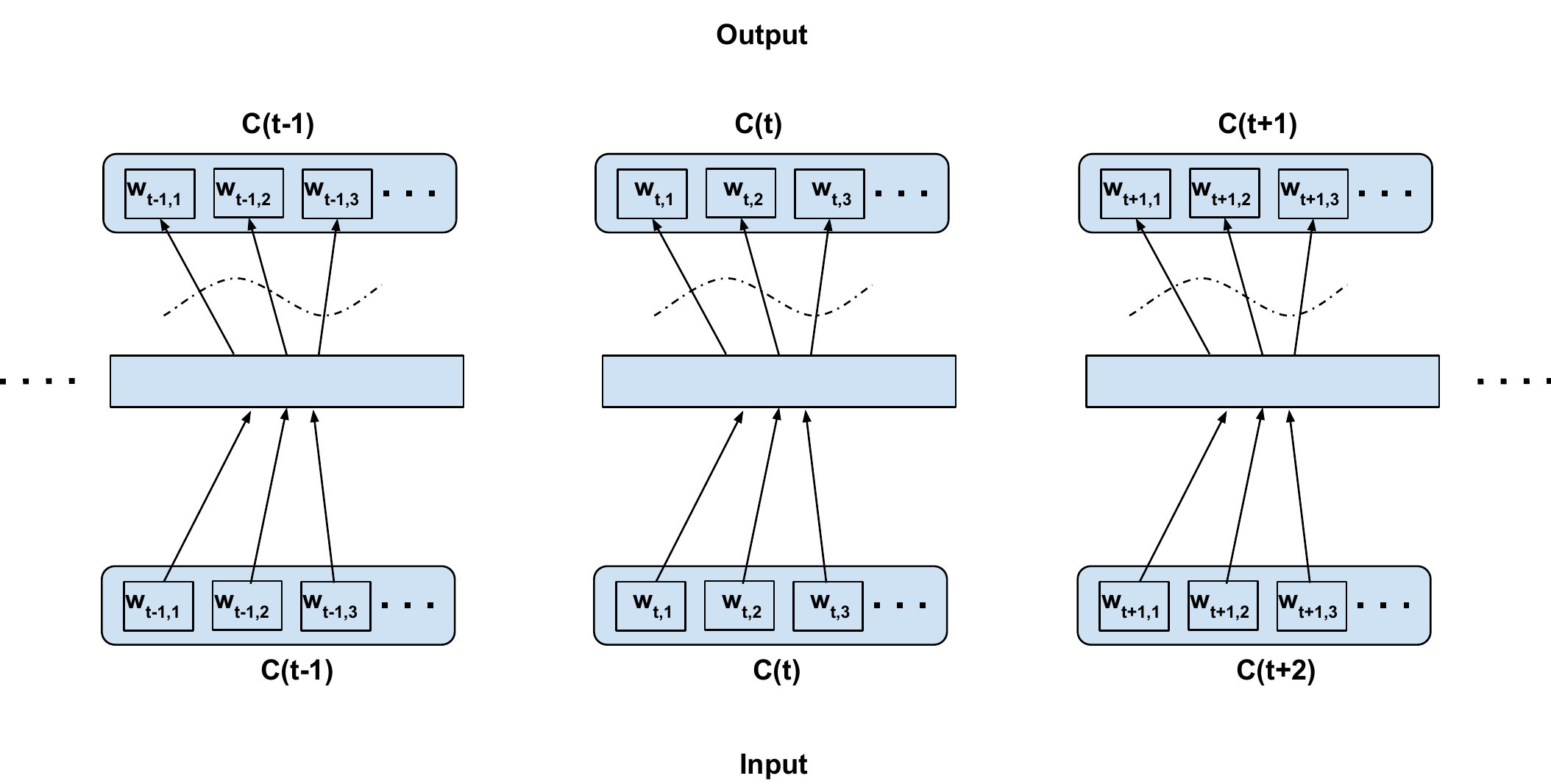}
	\captionsetup{justification=centering}
	\caption{\textbf{Proposed Intra-Confusion Training Scheme for Confusion networks.}\\
	\small{$c(t)$ is a unit word confusion in the confusion network at a time-stamp $t$, i.e., $c(t)$ represents a set of arcs between two adjacent nodes of a confusion network, representing a set of confusable words.\\
$w_{t,i}$ is the $i^{th}$ most probable word in the confusion $c(t)$.\\
	Word confusions are sorted in decreasing order of their posterior probability: $P(w_{t,1})>P(w_{t,2})>P(w_{t,3})...$\\
	The dotted curved lines denote that the self-mapping is disallowed.\\}}\label{fig:intra_confusion}
\end{figure*}

\subsection{Intra-Confusion Training - C2V-a}
Next, we explore the direct adaptation of the skip-gram modeling but on the confusion dimension (i.e., considering word confusions as contexts) rather than the traditional sequential context.
Figure~\ref{fig:intra_confusion} shows the training configuration over a confusion network.
In short every word is linked with every other alternate word in the confusion dimension (i.e., between set of confusable words) through the desired network (as opposed to the temporal context dimension in the word2vec training).
For clarity, since this is only using acoustically alternate words, we call this the C2V-acoustics or C2V-a model for short.
Note, we disallow any word being predicted from itself (this constrain is indicated with curved dotted lines in the figure).
As depicted in the Figure~\ref{fig:intra_confusion}, the word $w_{t,i}$ (confusion context) is predicted from $w_{t,j}$ (current word), where $i=1,2,3 \dots \text{length}(C(t))$ and $j \neq i$, for each $j=1,2,3 \dots \text{length}(C(t))$ for confusion $C(t)$ $\forall t$.
We expect such a model to capture inherent relations over the different word confusions.
In the context of an ASR lattice, we expect it to capture intrinsic relations between similarly sounding words (acoustically similar).
However, the model would fail to capture any semantic and syntactic relations associated with the language.
The embedding obtained from this configuration can be fused (concatenated) with the traditional skip-gram word2vec embedding to form a new subspace representing both the independently trained subspaces.
The number of training samples generated with this configuration is:
\begin{equation}\label{eq:intra}
\# \text{Samples} = \sum_{i=1}^{n} D_i \times (D_i - 1)
\end{equation}
where $n$ is the number of time steps, $D_i$ is the number of confusions at the $i^{th}$ time step.

\begin{figure*}[!b]
\centering
  \includegraphics[width=\linewidth]{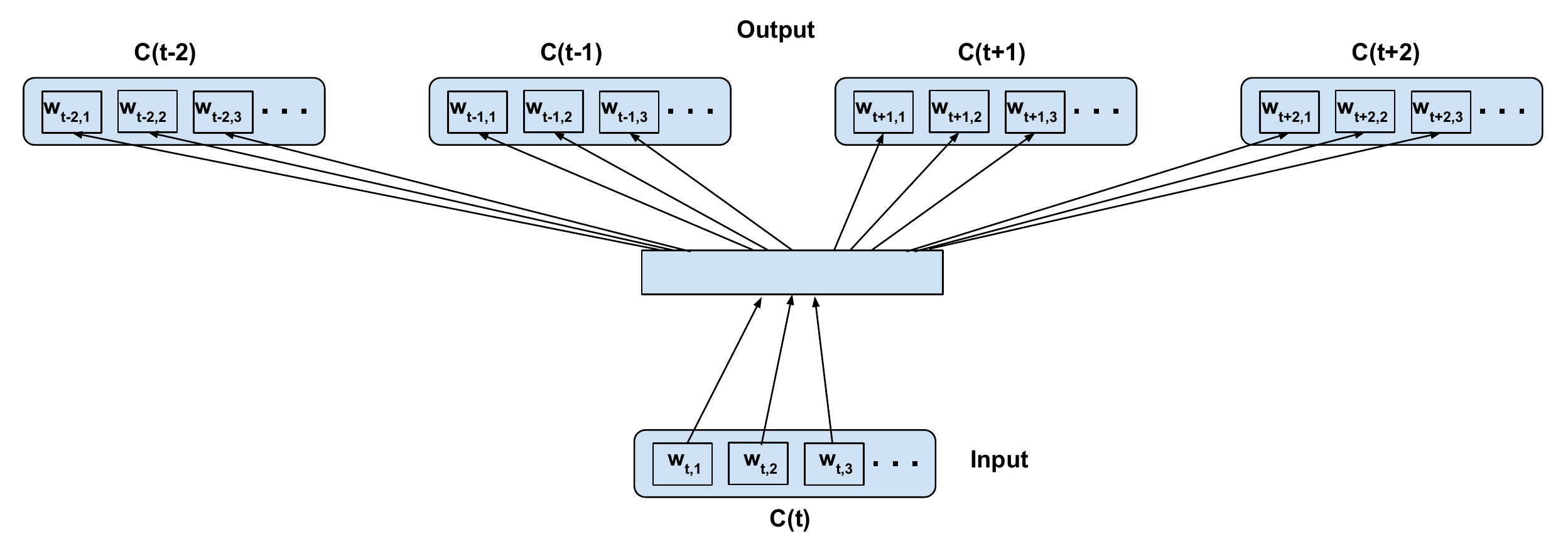}
	\captionsetup{justification=centering}
	\caption{\textbf{Proposed Inter-Confusion Training Scheme for Confusion networks.}\\
	\small{$c(t)$ is a unit word confusion in the confusion network at a time-stamp $t$, i.e., $c(t)$ represents a set of arcs between two adjacent nodes of a confusion network, representing a set of confusable words.\\
	$w_{t,i}$ is the $i^{th}$ most probable word in the confusion $c(t)$.
	\\Word confusions are sorted in decreasing order of their posterior probability: $P(w_{t,1})>P(w_{t,2})>P(w_{t,3})...$}}\label{fig:inter_confusion}
\end{figure*}

\subsection{Inter-Confusion Training - C2V-c}
In this configuration, we propose to model both the linguistic contexts and the word confusion contexts simultaneously.
Figure~\ref{fig:inter_confusion} illustrates the training configuration.
Each word in the current confusion is predicted from each word from the succeeding and preceding confusions over a predefined local context.
To elaborate, the words $w_{t-t',i}$ (context) are predicted from $w_{t,j}$ (current word) for $i=1,2,3 \dots \text{length}(C(t-t'))$, $j=1,2,3 \dots \text{length}(C(t))$, $t' \in {1,2,-1,-2}$ for skip-window of 2 for current confusion $C(t) \forall t$ as per Figure~\ref{fig:inter_confusion}.
Since we assume the acoustic similarities for a word to be co-occurring, we expect to jointly model the co-occurrence of both the context and confusions.
For clarity, since even the acoustic similarities are learned through context and not direct acoustic mapping, as in the Intra-confusion case, we call the inter-confusion training C2V-context or C2V-c for short.

This also has the additional benefit of generating more training samples than the intra-confusion training.
The number of training samples generated is given by:
\begin{equation}\label{eq:inter}
\# \text{Samples} = \sum_{i=1}^{n} \sum_{\substack{j=i-S_w \\ j\neq i}}^{i+S_w} D_i \times D_{j}
\end{equation}
where $n$ is the total number of time steps, $D_i$ is the number of word confusions at the $i^{th}$ time step, $S_w$ is the skip-window size (i.e., sample $S_w$ words from history and $S_w$ words from the future context of current word).
Inter-Confusion training can be viewed as an extension of top-confusion training where the skip-gram modeling is applied to all possible paths through the confusion network.

\begin{figure*}[t]
\centering
  \includegraphics[width=\linewidth]{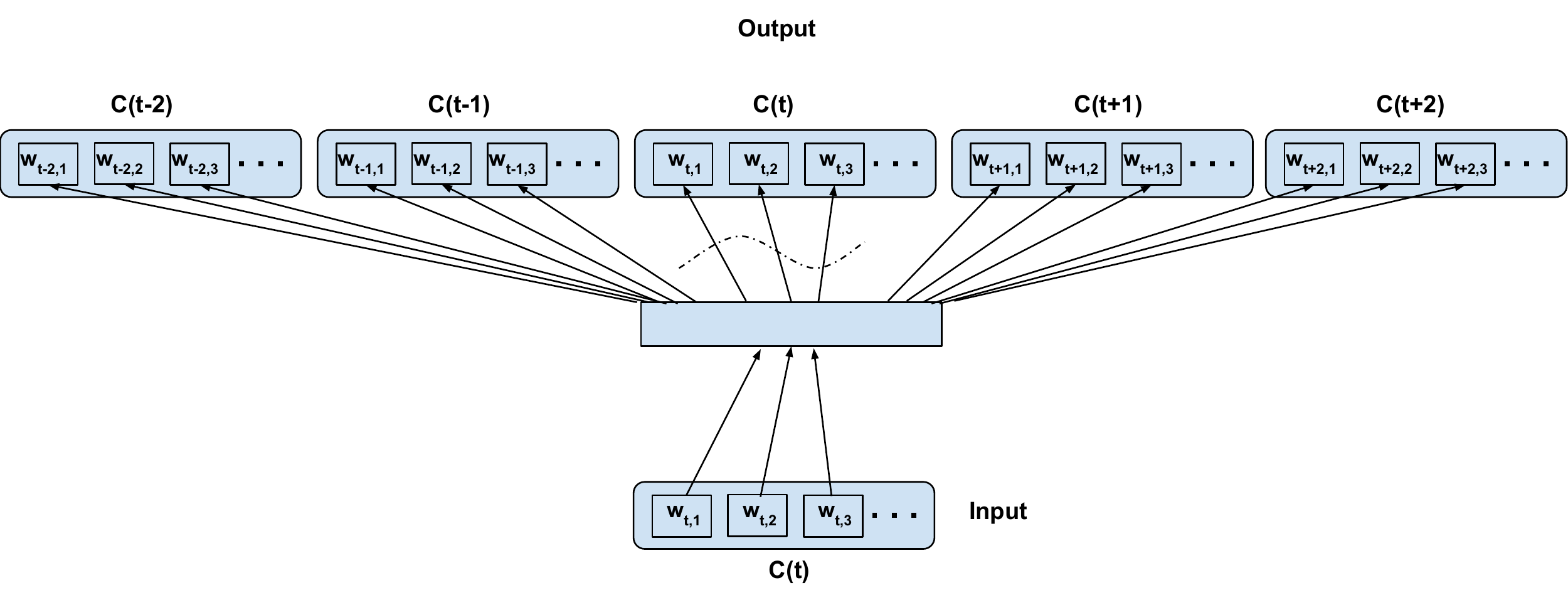}
	\captionsetup{justification=centering}
	\caption{\textbf{Proposed Hybrid-Confusion Training Scheme for Confusion networks.}\\
	\small{$c(t)$ is a unit word confusion in the confusion network at a time-stamp $t$, i.e., $c(t)$ represents a set of arcs between two adjacent nodes of a confusion network, representing a set of confusable words.\\
$w_{t,i}$ is the $i^{th}$ most probable word in the confusion $c(t)$.\\
Word confusions are sorted in decreasing order of their posterior probability: $P(w_{t,1})>P(w_{t,2})>P(w_{t,3})...$\\
The dotted curved lines denote that the self-mapping is disallowed.}}\label{fig:hybrid_confusion}
\end{figure*}

\subsection{Hybrid Intra-Inter Confusion Training - C2V-*}

Finally, we merge both the intra-confusion and inter-confusion training.
For clarity we call this model the C2V-* since it combines all the previous cases.
This can be seen as a super-set of top-confusion, inter-confusion and intra-confusion training configurations.
Figure~\ref{fig:hybrid_confusion} illustrates the training configuration.
The words $w_{t-t',i}$ (context) are predicted from $w_{t,j}$ (current word) for $i=1,2,3 \dots \text{length}(C(t-t'))$, $j=1,2,3 \dots \text{length}(C(t))$, $t' \in {1,2,0,-1,-2}$ such that if $t'=0$ then $i \neq j$; for skip-window of 2 for current confusion $C(t) \forall t$ as depicted in Figure~\ref{fig:hybrid_confusion}.
We simply add the combination of training samples from the above two proposed techniques (i.e., the number of samples is the sum of equation~\ref{eq:intra} and equation~\ref{eq:inter}).

\section{Training Schemes}\label{sec:tuning}

\begingroup
\subsection{Model Initialization/Pre-training}

Very often, it has been found that better model initializations lead to better model convergence \citep{erhan2010does}.
This is more significant in the case of under-represented words.
Moreover, for training the word confusion mappings, it would benefit to build upon the contextual word embeddings, since our final goal is in conjunction with both contextual and confusion information.
Hence, we experiment initializing all our models with the original Google's word2vec model\footnote{https://code.google.com/archive/p/word2vec/} trained on Google News dataset with 100 billion words as described by \cite{mikolov2013distributed}. Pre-training rules are explained in the flowchart in Figure~\ref{fig:tuning_schemes}(a).
For the words present in the Google's word2vec vocabulary, we directly initialize the embeddings with word2vec.
The embeddings for rest of the words are randomly initialized following uniform distribution.

\begin{figure*}[t]
\makebox[\linewidth][c]{
\centering
\includegraphics[width=\linewidth]{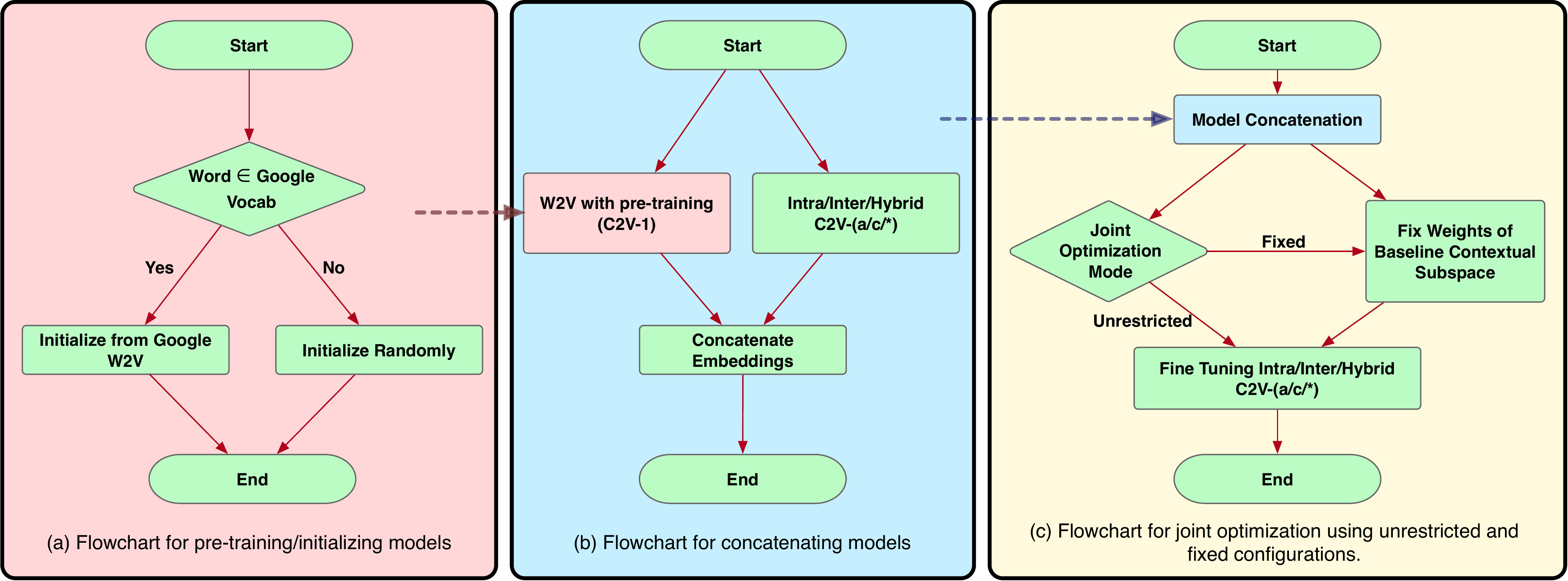}}%
\vspace{5mm}
\captionsetup{justification=centering}
\caption{\textbf{Flowcharts for proposed training schemes}}\label{fig:tuning_schemes}
\end{figure*}

\subsection{Model Concatenation}
The hypothesis with model concatenation is that the two subspaces, one representing the contextual subspace (word2vec), and the other capturing the confusion subspace can be both trained independently and concatenated to give a new vector space which manifests both the information and hence a potentially useful vector word representation. Flowchart for model concatenation is shown in Figure~\ref{fig:tuning_schemes}(b).
The model concatenation can be mathematically represented as:
\begin{equation}
NEW_{n \times e_1+e_2} = \begin{bmatrix} W2V_{n \times e_1} & C2V_{n \times e_2} \end{bmatrix}
\end{equation}

where $NEW$ is the new concatenated vector space of dimensions $n \times e_1 + e_2$, $n$ is the vocabulary size, $e_1$ and $e_2$ are the embedding sizes of $W2V$ and $C2V$ subspaces respectively.

\subsection{Joint Optimization}
Further to the model concatenation scheme, one could fine-tune the new vector space representation to better optimize to the task criterion (fine-tuning involves re-training end-to-end with a relatively lower learning rate than usual).
This could be viewed as a case of relaxing the strict independence between two subspaces as in the case of model concatenation.
The fine-tuning itself could be either of the aforementioned proposed techniques.
We specifically try two configurations of joint optimization:

\subsubsection{Fixed Contextual Subspace}
In this configuration, we fix the contextual (word2vec) subspace and fine-tune only the confusion subspace.
Since the word2vec already provides robust contextual representation, any fine-tuning on contextual space could possibly lead to sub-optimal state.
Keeping the word2vec subspace fixed also allows the model to concentrate more specifically towards the confusion since the fixed subspace compensates for all the contextual mappings during training.
This allows us to constrain the updatable parameters during joint optimization.
It also allows for the possibility to directly use available word2vec models without modifications.
The flowchart for the Fixed Contextual Subspace joint optimization is displayed in Figure~\ref{fig:tuning_schemes}(c).
\endgroup

\subsubsection{Unrestricted}
In this configuration, we optimize both the subspaces, i.e., the contextual (word2vec) and the confusion subspaces.
The hypothesis is the fine-tuning allows the two subspaces to interact to achieve the best possible representation.
The flowchart for the unrestricted joint optimization is displayed in Figure~\ref{fig:tuning_schemes}(c).

\section{Evaluation Methods}\label{sec:evaluation}
Prior literature suggests, there are two prominent ways for evaluating the vector space representation of words.
One is based on Semantic\&Syntactic analogy task as introduced by \cite{mikolov2013efficient}.
The other common approach has been to assess the word similarities by computing the rank-correlation (Spearman's correlation) on human annotated word similarity databases \citep{schnabel2015evaluation} like WordSim-353 \citep{finkelstein2001placing}.
Although, the two evaluations can judge the vector representations of words efficiently for semantics and syntax of a language, we need to device an evaluation criteria for the word confusions, specifically for our case scenario - the acoustic confusions of words.
For this, we formulate evaluations for acoustic confusions parallel to the analogy task and the word similarity task.

\subsection{Analogy Tasks}\label{sec:evaluation_analogy_tasks}
\subsubsection{Semantic\&Syntactic Analogy Task}
\cite{mikolov2013efficient} introduced an analogy task for evaluating the vector space representation of words. 
The task was based on the intuition that the words, say ``king'' is similar to ``man'' in the same sense as the ``queen'' is to ``woman'' and thus relies on answering questions relating to such analogies by performing algebraic operations on word representations. 
For example, the analogy is correct if the vector(``woman'') is most similar to vector(``king'')-vector(``man'')+vector(``queen'').
The analogy question test set is designed to test both syntactic and semantic word relationships.
The test set contains five types of semantic questions (8869 questions) and nine types of syntactic questions (10675 questions).
Finally, the efficiency of the vector representation is measured using the accuracy achieved on the analogy test set.
We employ this for testing the Semantic \& Syntactic (contextual axis as in terms of Figure~\ref{fig:asr_sausage}) relationship inherent in the vector space.

\subsubsection{Acoustic Analogy Task}
\begin{table}[t]
\begin{center}
\setlength{\tabcolsep}{15pt}
\begin{tabular}{|c|c||c|c|}
\hline
\multicolumn{2}{|c||}{Word Pair 1} & \multicolumn{2}{|c|}{Word Pair 2} \\
\hline
i'd & eyed & phi & fie \\
seedar & cedar & rued & rude \\
air & aire & spade & spayed \\
scent & cent & vile & vial \\
cirrus & cirrous & sold & soled \\
curser & cursor & pendant & pendent \\
sensor & censor & straight & strait \\
\hline
\end{tabular}
\end{center}
\caption{\textbf{Few examples from Acoustic Analogy Task Test-set}}\label{tab:conf_analogy}
\end{table}

The primary purpose of the acoustic analogy task is to independently gauge the acoustic similarity information captured by the embedding model irrespective of the inherent Semantic and Syntactic linguistic information.
Adopting similar idea and extending the same for evaluation of word confusions, we formulate the acoustic confusion analogy task (vertical context test as in terms of Figure~\ref{fig:asr_sausage}) as follows.
For similar sounding word pairs, ``see'' \& ``sea'' and ``red'' \& ``read'', the word vector ``see'' is similar to ``sea'' in the same sense as the word ``red'' is to ``read''.
We set up an acoustic analogy question set on acoustically similar sounding words, more specifically homophones.
Table~\ref{tab:conf_analogy} lists a few examples from our data set.
Detailed description of the creation of dataset is presented in section~\ref{sec:data_acoustic_task}.

\begin{table}[t]
\begin{center}
\resizebox{\linewidth}{!}{%
\begin{tabular}{|c||c|c||c|c|}
\hline
Type of Relationship & \multicolumn{2}{|c||}{Word Pair 1} & \multicolumn{2}{|c|}{Word Pair 2} \\
\hline
\multirow{3}{*}{Currency} & India & Rupee & Korea & One (Won) \\
& Canada & Dollar & Denmark & Krona (Krone) \\
& Japan & Yen & Sweden & Krone (Krona) \\
\hline
\multirow{3}{*}{Family} & Buoy (Boy) & Girl & Brother & Sister \\
& Boy & Girl & King & Quean (Queen) \\
& Boy & Girl & Sun (Son) & Daughter \\
\hline
Adjective-to-Adverb & Calm & Calmly & Sloe (Slow) & Slowly \\
\hline
Opposite & Aware & Unaware & Possible & Impassible (Impossible) \\
\hline
Comparative & Bad & Worse & High & Hire (Higher) \\
\hline
Superlative & Bad & Worst & Grate (Great) & Greatest \\
\hline
Present Participle & Dance & Dancing & Rite (Write) & Writing \\
\hline
Past Tense & Dancing & Danced & Flying & Flu (Flew) \\
\hline
Plural & Banana & Bananas & Burred (Bird) & Birds \\
\hline
Plural Verbs & Decrease & Decreases & Fined (Find) & Finds \\
\hline
\hline
\multirow{3}{*}{Multiple Homophone Substitutions} & Wright (Write) & Writes & Sea (See) & Sees \\
& Rowed (Road) & Roads & I (Eye) & Ayes (Eyes) \\
& Si (See) & Seize (Sees) & Right (Write) & Writes \\
\hline
\end{tabular}}
\end{center}
\captionsetup{justification=centering}
\caption{\textbf{Few examples from Semantic \& Syntactic - Acoustic Analogy Task Test Set}\\
\small{The words in the parenthesis are the original ones as in the analogy test set \citep{mikolov2013efficient} which have been replaced by their homophone alternatives.}}\label{tab:lang_conf_analogy}
\end{table}

\subsubsection{Semantic\&Syntactic-Acoustic Analogy Task}

Further, rather than evaluating the Semantic-Syntactic tasks and the acoustic analogy tasks independently, we could test for both together.
Intuitively, the word vectors in each of the two subspaces should interact together.
We would expect for an analogy, ``see''-``saw'':``take''-``took'', the word ``see'' has a homophone alternative in ``sea'', thus there is a possibility of the word ``see'' being confused with ``sea'' in the new vector space.
Thus an algebraic operation such as $vector(``see")-vector(``saw")+vector(``take")$ should be similar to $vector(``took")$ as before.
Moreover the $vector(``sea")-vector(``saw")+vector(``take")$ should also be similar to $vector(``took")$.
This is because we expect the $vector(``sea")$ to be similar to $vector(``see")$ under the acoustic subspace.
We also take into account the more challenging possibility of more than one homophone word substitution.
For example, $vector(``see")-vector(``saw")+vector(``allow")$ is similar to $vector(``allowed")$, $vector(``aloud")$ and $vector(``sea")-vector(``saw")+vector(``allow")$.
The hypothesis is that to come up with such a representation the system should jointly model both the language semantic-syntactic relations and the acoustic word similarity relations between words.
The task is designed to test Semantic-Acoustic relations and the Syntactic-Acoustic relationships.
In other words, in terms of Figure~\ref{fig:asr_sausage}, the task evaluates both the horizontal \& vertical context together.
A few examples of this task is listed in Table~\ref{tab:lang_conf_analogy}.
Section~\ref{sec:data_ss_acoustic_task} details the creation of the database.

\subsection{Similarity Ratings}
\subsubsection{Word Similarity Ratings}
Along with the analogy task the word similarity task \citep{finkelstein2001placing} has been popular to evaluate the quality of word vector representations in the NLP community \citep{pennington2014glove,luong2013better,huang2012improving,schnabel2015evaluation}.
In this work we employ the WordSim-353 dataset \citep{finkelstein2001placing} for the word similarity task.
The dataset has a set of 353 word pairs with diverse range of human annotated scores relating to the similarity/dissimilarity of the two words.
The rank-order correlation (Spearman correlation) between the human annotated scores and the cosine similarity of word vectors is computed.
Higher correlation corresponds to better preservation of word similarity order represented by the word vectors, and hence better quality of the embedding vector space.

\subsubsection{Acoustic Similarity Ratings}
\begin{table}[!b]
\begin{center}
\begin{tabular}{|c|c|c|c|}
\hline
Word1 & Word2 & Acoustic Rating & WordSim353\\
\hline
I & Eye & 1.0 & - \\
Adolescence & Adolescents & 0.9 & -\\
Allusion & Illusion & 0.83 & - \\
Sewer & Sower & 0.66 & - \\
Fighting & Defeating & 0.57 & 7.41 \\
Day & Dawn & 0.33 & 7.53 \\
Weather & Forecast & 0.0 & 8.34\\
\hline
\end{tabular}
\end{center}
\captionsetup{justification=centering}
\caption{\textbf{Examples of Acoustic Similarity Ratings}\\
\small{Acoustic Rating: 1.0 = Identically sounding, 0.0 = Highly acoustically dissimilar\\
WordSim353: 10.0 = High word similarity, 0.0 = Low word similarity\\
Word pairs not present in WordSim353 is denoted by '-'}}\label{tab:acoustic_conf_ratings}
\end{table}
Employing a similar analogous idea to word similarity ratings and extending it to reflect the quality of word confusions, we formulate an acoustic word similarity task.
The attempt is to have word pairs scored similar to as in WordSim-353 database, but with the scores reflecting the acoustic similarity.
Table~\ref{tab:acoustic_conf_ratings} lists a few randomly picked examples from our dataset.
The dataset generation is described in section~\ref{sec:data_acoustic_sim}.

\section{Data \& Experimental Setup}\label{sec:baseline_setup}
\subsection{Database}
We employ Fisher English Training Part 1, Speech (LDC2004S13) and Fisher English Training Part 2, Speech (LDC2005S13) corpora \citep{cieri2004fisher} for training the ASR.
The corpora consists of approximately 1915 hours of telephone conversational speech data sampled at 8kHz.
A total of 11972 speakers were involved in the recordings.
The speech corpora is split into three speaker disjoint subsets for training, development and testing for ASR modeling purposes.
A subset of the speech data containing approximately 1905 hours were segmented into 1871731 utterances to train the ASR.
Both the development set and the test set consists of 5000 utterance worth 5 hours of speech data each.
The transcripts contain approximately 20.8 million word tokens with 42150 unique entries.

\subsection{Experimental Setup}

\subsubsection{Automatic Speech Recognition}
KALDI toolkit is employed for training the ASR \citep{povey2011kaldi}.
A hybrid DNN-HMM based acoustic model is trained on high resolution (40 dimensional) Mel Frequency Cepstral Coefficients (MFCC) along with i-vector features to provide speaker and channel information for robust modeling.
The CMU pronunciation dictionary \citep{weide1998cmu} is pruned to corpora's vocabulary and is used as a lexicon for the ASR.
A trigram language model is trained on the transcripts of the training subset data.
The ASR system achieves a word error rates (WER) of 16.57\% on the development and 18.12\% on the test datasets.
The decoded lattice is used to generate confusion network based on minimum bayes risk criterion \citep{xu2011minimum}.
The ASR output transcriptions resulted in a vocabulary size of 41274 unique word tokens.

\subsubsection{Confusion2Vec}
For training the Confusion2Vec, the training subset of the Fisher corpora is used.
The total number of tokens resulting from the multiple paths over the confusion network is approximately 69.5 million words, i.e., an average of 3.34 alternative word confusions present for each word in the confusion network.
A minimum frequency threshold of 5 is set to prune the rarely occurring tokens from the vocabulary, which resulted in the reduction of the vocabulary size from 41274 to 32848.
Further, we also subsample the word tokens as suggested by \cite{mikolov2013distributed} which was shown to be helpful.
Both the frequency thresholding and the downsampling resulted in a reduction of word tokens from 69.5 million words to approximately 33.9 million words.
The Confusion2Vec and Word2Vec are trained using the Tensorflow toolkit \citep{abadi2016tensorflow}.
Negative Sampling objective is used for training as suggested for better efficiency \citep{mikolov2013distributed}.
For the skip-gram training, the batch-size of 256 was chosen and 64 negative samples were used for computing the negative sampling loss.
The skip-window was set to 4 and was trained for a total of 15 epochs.
The parameters were chosen to provide optimal performance with traditional word2vec embeddings, evaluating for word analogy task, for the size of our database.
During fine-tuning, the model was trained with a reduced learning rate and with other parameters unchanged.
All the above parameters were fixed for consistent and fair comparison.

\subsection{Creation of Evaluation Datasets}

\subsubsection{Acoustic Analogy Task}\label{sec:data_acoustic_task}
We collected a list of homophones in English \footnote{http://homophonelist.com/homophones-list/ (Accessed: 2018-04-30)}, and created all possible combinations of pairs of acoustic confusion analogies.
For homophones with more than 2 words, we list all possible confusion pairs.
Few examples from the dataset are listed in Table~\ref{tab:conf_analogy}.
We emphasize that the consideration of only homophones in the creation of the dataset is a strict and a difficult task to solve, since the ASR lattice contains more relaxed word confusions.

\subsubsection{Semantic\&Syntactic-Acoustic Analogy Task}\label{sec:data_ss_acoustic_task}
We construct an analogy question test set by substituting the words in the original analogy question test set from \cite{mikolov2013efficient} with their respective homophones.
Considering all the 5 types of semantic questions and 9 types of syntactic questions, for any words in the analogies with homophone alternatives, we swap with the homophone.
We prune all the original analogy questions having no words with homophone alternatives.
For analogies having more than one words with homophone alternatives, we list all permutations.
We found that the number of questions generating by the above method, being exhaustive, was large and hence we randomly sample from the list to retain 948 semantic questions and 6586 syntactic questions.
Table~\ref{tab:lang_conf_analogy} lists a few examples with single and multiple homophone substitutions for Semantic\&Syntactic-Acoustic Analogy Task from our data set.

\subsubsection{Acoustic Similarity Task}\label{sec:data_acoustic_sim}
\begin{table}[t]
\begin{center}
\begin{tabular}{|c|c|c|}
\hline
Task & Total Samples & Retained \\
\hline
Semantic\&Syntactic Analogy & 19544 & 11409 \\
Acoustic Analogy & 20000 & 2678 \\
Semantic\&Syntactic-Acoustic Analogy & 7534 & 3860 \\
WordSim-353 & 353 & 330 \\
Acoustic Confusion Ratings & 1372 & 943 \\
\hline
\end{tabular}
\end{center}
\caption{\textbf{Statistics of Evaluation Datasets}}\label{tab:stats_eval_data}
\end{table}
To create a set of word pairs scored by their acoustic similarity, we add all the homophone word pairs with an acoustic similarity score of $1.0$.
To get a more diverse range of acoustic similarity scores, we also utilize all the 353 word pairs from the WordSim-353 dataset and compute the normalized phone edit distance using the CMU Pronunciation Dictionary \citep{weide1998cmu}.
The normalized phone edit distance is of the range between $0$ and $1$.
The edit distance of $1$ refers to the word pair having almost $0$ overlap between their respective phonetic transcriptions and thus being completely acoustically dissimilar and vice-versa.
We use $1-\text{phone-edit-distance}$ as the acoustic similarity score between the word pair.
Thus a score of $1.0$ signifies that the two words are identically sounding, whereas as $0$ refers to words sounding drastically dissimilar.
In the case of a word having more than one phonetic transcriptions (pronunciation alternatives), we use the minimum normalized edit distance.
Table~\ref{tab:acoustic_conf_ratings} shows a few randomly picked examples from the generated dataset.
\vspace{3mm}\newline Finally, for evaluation the respective corpora are pruned to match the in-domain training dataset vocabulary. Table~\ref{tab:stats_eval_data} lists the samples in each evaluation dataset before and after pruning.

\subsection{Performance Evaluation Criterion}

In the original work by \cite{mikolov2013efficient}, the efficiency of the vector representation is measured using the accuracy achieved on the analogy test set.
But, in our case, note that the Semantic\&Syntactic analogy task and the Semantic\&Syntactic-Acoustic analogy task are mutually exclusive of each other.
In other words, the model can get only one, either one, of the analogies correct meaning any increments with one task will result in decrements over the other task.
Moreover, while jointly modeling two orthogonal information streams (i) contextual co-occurrences, and (ii) acoustic word confusions, finding the nearest word vector nearest to the specific analogy is no longer an optimal evaluation strategy.
This is because the word vector nearest to the analogy operation can either be along the contextual axis or the confusion axis, i.e., each analogy could possibly have two correct answers.
For example, the analogy ``write''-``wrote'' : ``read'' can be right when the nearest word vector is either ``read'' (contextual dimension) or ``red'' (confusion dimension).
To incorporate this, we provide the accuracy over top-2 nearest vectors, i.e., we count the analogy question as correct if any of the top-2 nearest vector satisfies the analogy.
This also holds for the acoustic confusion analogy tasks, especially for relations involving triplet homophones.
For example, the analogy ``write'' - ``right'' : ``road'' can be right when the nearest word vector is either ``rode'' or ``rowed'' (for triplet homophones ``road''/``rode''/``rowed'').
Thus, we present evaluations by comparing the top-1 (nearest vector) evaluation with baseline word2vec against the top-2 evaluation for the proposed confusion2vec models. %
To maintain consistency, we also provide the top-2 evaluations for the baseline word2vec models in the appendix.

Moreover, since we have 3 different analogy tasks, we provide the average accuracy among the 3 tasks in order to have an easy assessment of the performance of various proposed models.

\section{Results}\label{sec:results}
\begin{table*}[t]
\begin{center}
\resizebox{\textwidth}{!}{%
\begin{tabular}{|l|c|c|c|c|c|c|}
\hline
\multirow{2}{*}{Model} & \multicolumn{4}{|c|}{Analogy Tasks} & \multicolumn{2}{|c|}{Similarity Tasks} \\
\cline{2-7}
& S\&S & Acoustic & S\&S-Acoustic & Average Accuracy & Word Similarity & Acoustic Similarity \\
\hline
Google W2V & \textbf{61.42\%} & 0.9\% & 16.99\% & 26.44\% & \textbf{0.6893} & -0.3489 \\
In-domain W2V & 35.15\% & 0.3\% & 7.86\% & 14.44\% & 0.5794 & -0.2444 \\
C2V-1 & 43.33\% & 1.16\% & 15.05\% & 19.85\% & 0.4992 & 0.1944 \\
C2V-a & 22.03\% & 52.58\% & 14.61\% & 29.74\% & $0.105^{\ast}$ & \textbf{0.8138} \\
C2V-c & 36.15\% & \textbf{60.57\%} & 20.44\% & \textbf{39.05\%} & 0.2937 & 0.8055 \\
C2V-* & 30.53\% & 53.55\% & \textbf{29.35\%} & 37.81\% & $0.0963^{\ast}$ & 0.7858 \\
\hline
\end{tabular}}
\end{center}
\captionsetup{justification=centering}
\caption{\textbf{Results: Different proposed models}\\
\small{\textbf{C2V-1: Top-Confusion, C2V-a: Intra-Confusion, C2V-c: Inter-Confusion, C2V-*: Hybrid Intra-Inter}\\
All the models are of 256 dimensions except Google's W2V which is 300 dimensions.\\
For the analogy tasks: the accuracies of baseline word2vec models are for top-1 evaluations, whereas of the other models are for top-2 evaluations (as discussed in Section~\ref{sec:evaluation_analogy_tasks}).
Detailed semantic analogy and syntactic analogy accuracies, the top-1 evaluations and top-2 evaluations for all the models are available under Appendix in Table~\ref{appendix:tab:results}.\\
For the similarity tasks: all the correlations (Spearman's) are statistically significant with $p<0.001$ except the ones with the asterisks.
Detailed $p-values$ for the correlations are presented under Appendix in Table~\ref{appendix:tab:results:similarity}.\\
S\&S: Semantic \& Syntactic Analogy.}}\label{tab:results}
\end{table*}

Table ~\ref{tab:results} lists the results for various models.
We provide evaluations on three different analogy tasks and two similarity tasks as discussed in Section~\ref{sec:evaluation}.
Further, more thorough results with the Semantic and Syntactic accuracy splits are provided under the appendix to gain deeper insights.

\subsection{Baseline Word2Vec Model}
We consider 2 variations of Word2Vec baseline model.
First, we provide results with the Google's Word2Vec model \footnote{https://code.google.com/archive/p/word2vec} which is trained with orders more training data, and is thus a high upper bound on the Semantic\&Syntactic task.
The Google's Word2Vec model was pruned to match the vocabulary of our corpora to make the evaluation comparable.
Second, we consider the Word2Vec model trained on the in-domain ground truth transcripts.
The two baseline models result in good performance on Semantic\&Syntactic analogy tasks and word similarity task as expected.
The Google's model achieves an accuracy of 61.42\% on the Semantic\&Syntactic analogy task.
We note that the Syntactic accuracy (70.79\%) is much higher than the Semantic accuracy (28.98\%) (see Appendix Table~\ref{appendix:tab:results}).
This could be due to our pruned evaluation test set (see Table~\ref{tab:stats_eval_data}).
The in-domain model improves on the Semantic accuracy while losing on the syntactic accuracy over the Google model (see Appendix Table~\ref{appendix:tab:results}).
The shortcomings of the in-domain model compared to the Google Word2Vec on the Semantic\&Syntactic analogy task can be attributed towards the amount of training data and its extensive vocabulary.
The in-domain model is trained on 20.8 million words versus the 100 billion of Google's News dataset. 
Moreover the vocabulary size of in-domain models is approximately 42,150 versus the 3 million of Google \citep{mikolov2013distributed} and thus unfair to compare with rest of the models.
Further, evaluating the Acoustic analogy and Semantic\&Syntactic-Acoustic analogy tasks, all the baseline models perform poorly.
An unusual thing we note is that the Google Word2Vec model performs better comparatively to the in-domain baseline model in the Semantic\&Syntactic-Acoustic analogy task.
A deeper examination revealed that the model compensates well for homophone substitutions on Semantic\&Syntactic analogies which have very similar spellings.
This suggests that the typographical errors present in the training data of the Google model results in a small peak in performance for the Semantic\&Syntactic-Acoustic analogy task.
On the evaluations of similarity tasks, all the baseline models perform well on the word similarity tasks as expected.
However, they exhibit poor results on the acoustic similarity task. 
Overall, the results indicate that the baseline models are largely inept of capturing any relationships over the acoustic word confusions present in a confusion network or a lattice.
In our specific case, the baseline models are poor in capturing relationships between acoustically similar words.

\subsection{Top-Confusion - C2V-1}
Comparing the top-confusion (C2V-1 for short), training scheme with the baseline in-domain word2vec model, we observe the baseline model trained on clean data performs better on the Semantic\&Syntactic analogy task as expected.
Baseline in-domain word2vec achieves 35.15\% on the Semantic\&Syntactic analogy task whereas the top-confusion model achieves 34.27\% (see Appendix Table~\ref{appendix:tab:results}).
However, the performance difference is minimal.
This is encouraging because the top-confusion model is trained on the noisy ASR transcripts.
Moreover, we see the noisy transcripts negatively affect the semantic accuracies while the syntactic accuracy remains identical which makes sense (see Appendix Table~\ref{appendix:tab:results}).
Similar to the baseline in-domain word2vec model, the top-confusion model falls short to Google word2vec mainly due to the extensive amount of data employed in the latter case.

Evaluating for Acoustic analogies and Semantic\&Syntactic-Acoustic analogies, the top-confusion scheme improves slightly over the baseline word2vec model. 
This hints at the ability of the top-confusion model to capture some acoustic word confusions through context (e.g. ``take a seat'' is expected but sometimes we may see ``take a sit'').
The improvements are small because in a good quality ASR the top confusion network hypotheses contain few errors thus context learning is much stronger and acoustic-confusion learning is minimal.
Note that the top-confusion model would converge to the baseline word2vec model in the case of zero word error rate.

Further, inspecting the performance in the similarity task, the top-confusion model exhibits statistically significant positive correlation in the word similarity task, although slightly smaller correlation than the baseline word2vec and Google word2vec model.
However, we observe a positive (statistically significant) correlation on the acoustic similarity task, whereas both the baseline word2vec and Google word2vec model exhibit a negative correlation.
This further validates the proposed top-confusion model's capability to capture acoustic word confusions.

\subsection{Intra-Confusion, C2V-a}
With intra-confusion training (C2V-acoustic or C2V-a for short) we expect the model to capture acoustically similar word relationships, while completely ignoring any contextual relations.
Hence, we expect the model to perform well on acoustic analogies and acoustic similarity tasks and to perform poorly on Semantic\&Syntactic analogies and word similarity tasks.
The Table~\ref{tab:results} lists the results obtained using intra-confusion training.
The results are in conjunction with our expectations.
The model gives the worst results in Semantic\&Syntactic analogy task.
However, we observe that the syntactic analogy accuracy to be a fair amount higher than the semantic accuracy (see Appendix Table~\ref{appendix:tab:results}).
We think this is mainly because of syntactically similar words appearing along the word confusion dimension in the confusion networks, resultant of the constraints enforced on the confusion network by the (ASR) language model - which are known to perform better for syntactic tasks \citep{mikolov2013efficient}.
The model also gives the highest correlation on the acoustic similarity task, while performing poorly on the word similarity task.

\subsection{Inter-Confusion, C2V-c}
With inter-confusion training (C2V-contextual or C2V-c for short), we hypothesized that the model is capable of jointly modeling both the contextual information as well as confusions appearing contextually.
Hence, we expect the model to perform well on both the Semantic\&Syntactic analogy and Acoustic analogy tasks and in doing so result in better performance with Semantic\&Syntactic-Acoustic analogy task.
We also expect the model to give high correlations for both word similarity and acoustic similarity tasks.
From Table~\ref{tab:results}, we observe that as hypothesized the inter-confusion training shows improvements in the Semantic\&Syntactic analogy task.
Quite surprisingly, the inter-confusion training shows better performance than the intra-confusion training for the Acoustic analogy task, hinting that having good contextual representation could mutually be beneficial for the confusion representation.
However, we don't observe any improvements in the Semantic\&Syntactic-Acoustic analogy task.
Evaluating on the similarity tasks, the results support the observations drawn from analogy tasks, i.e., the model fares relatively well in both word similarity and acoustic similarity.

\subsection{Hybrid Intra-Inter Confusion, C2V-*}
The hybrid intra-inter confusion training (C2V-* for short) introduces all confusability and allows learning directly confusable acoustic terms, such as in the C2V-a case, and contextual information that incorporates confusable terms, as in the Inter or C2V-c case.
This model shows comparable performance in jointly modeling on both the Semantic\&Syntactic and Acoustic analogy tasks.
One crucial observation is that it gives significantly better performance with the Semantic\&Syntactic-Acoustic analogy task.
This suggests that jointly modeling both the intra-confusion and inter-confusion word mappings is useful.
However, it achieves better results by compromising on the semantic analogy (see Appendix Table~\ref{appendix:tab:results}) accuracy and hence also negatively affecting the word similarity task.
The model achieves good correlation on the acoustic similarity task.\\

\begin{table*}[t]
\begin{center}
\resizebox{\textwidth}{!}{%
\begin{tabular}{|l|c|c|c|c|c|c|}
\hline
\multirow{2}{*}{Model} & \multicolumn{4}{|c|}{Analogy Tasks} & \multicolumn{2}{|c|}{Similarity Tasks} \\
\cline{2-7}
& S\&S & Acoustic & S\&S-Acoustic & Average Accuracy & Word Similarity & Acoustic Similarity \\
\hline
Google W2V & \textbf{61.42\%} & 0.9\% & 16.99\% & 26.44\% & \textbf{0.6893} & -0.3489 \\
In-domain W2V & 59.17\% & 0.6\% & 8.15\% & 22.64\% & 0.4417 & -0.4377 \\
C2V-1 & 61.13\% & 0.9\% & 16.66\% & 26.23\% & \textbf{0.6036} & -0.4327 \\
C2V-a & 63.97\% & 16.92\% & \textbf{43.34\%} & 41.41\% & 0.5228 & 0.62 \\
C2V-c & \textbf{65.45\%} & \textbf{27.33\%} & 38.29\% & \textbf{43.69\%} & 0.5798 & 0.5825 \\
C2V-* & 65.19\% & 20.35\% & 42.18\% & 42.57\% & 0.5341 & \textbf{0.6237} \\
\hline
\end{tabular}%
}
\end{center}
\captionsetup{justification=centering}
\caption{\textbf{Results with pre-training/initialization}\\
\small{\textbf{C2V-1: Top-Confusion, C2V-a: Intra-Confusion, C2V-c: Inter-Confusion, C2V-*: Hybrid Intra-Inter}\\
All the models are of 300 dimensions.\\
For the analogy tasks: the accuracies of baseline word2vec models are for top-1 evaluations, whereas of the other models are for top-2 evaluations (as discussed in Section~\ref{sec:evaluation_analogy_tasks}).
Detailed semantic analogy and syntactic analogy accuracies, the top-1 evaluations and top-2 evaluations for all the models are available under Appendix in Table~\ref{appendix:tab:pretraining_results}.\\
For the similarity tasks: all the correlations (Spearman's) are statistically significant.
Detailed $p-values$ for the correlations are presented under Appendix in Table~\ref{appendix:tab:pretraining_results:similarity}.\\
S\&S: Semantic \& Syntactic Analogy.}}\label{tab:pretraining_results}
\end{table*}

\Pg{Summary Model Comparison}
Overall, our proposed Confusion2Vec models capture significantly more useful information compared to the baseline models judging by the average accuracy over the analogy tasks.
One particular observation we see across all the proposed models is that the performance remains fairly poor for the Semantic\&Syntactic-Acoustic analogy task.
This suggests that the Semantic\&Syntactic-Acoustic analogy task is inherently hard to solve.
We believe that to achieve better results with Semantic\&Syntactic-Acoustic analogies, it is necessary to have robust performance on one of the tasks (Semantic\&Syntactic analogies or Acoustic analogies) to begin with, i.e., better model initialization could help.
Next, we experiment with model initializations/pre-training.

\subsection{Model Initialization/Pre-training}

Table~\ref{tab:pretraining_results} lists the results with model initialization/pre-training.
The in-domain word2vec baseline model and the top-confusion models are initialized from the Google Word2Vec model.
Pre-training the models provide improvements with Semantic\&Syntactic analogy results to be close and comparable to that of the Google's Word2Vec model.
Empirically, we find the top-confusion model inherits approximately similar contextual information as the baseline models, and in addition outperforms the baseline in average accuracy.
Thus, for the future experiments we adopt top-confusion model (rather than word2vec model) for initialization, model concatenation and joint-training.
The remaining models (C2V-a, C2V-c, and C2V-*) are initialized from the top-confusion model (i.e., C2V-1, the top-confusion model initialized from Google Word2Vec), since this would enable full compatibility with the vocabulary.
Since the Google Word2Vec model is 300 dimensional, this forces all the pre-trained models (in Table~\ref{tab:pretraining_results}) to be 300, opposed to 256 dimensions (in Table~\ref{tab:results}).

For intra-confusion model, the pre-training provides drastic improvements on Semantic\&Syntactic analogy task at the expense of the Acoustic analogy task.
Even-though the accuracy of Acoustic analogy task decreases comparatively to without pre-training, it remains significantly better than the baseline model.
More importantly, the Semantic\&Syntactic-Acoustic analogy task accuracy doubles.
Inter-Confusion model does not compromise on the Semantic\&Syntactic analogy tasks, in doing so gives comparable performances to the baseline model.
Additionally it also does well on the Acoustic and Semantic\&Syntactic-Acoustic analogy task as was the case without pre-training.
In the case of hybrid intra-inter confusion model, similar trends are observed as was with no pre-training, but with considerable improvements in accuracies.
Pre-training also helps in boosting the correlations for the word similarity tasks for all the models.
Overall, we find the pre-training to be extremely useful.

\begin{table*}[t]
\begin{center}
\resizebox{\linewidth}{!}{%
\begin{tabular}{@{\makebox[1.75em][r]{\rownumber\space}}|c|c|c|c|c|c|c|c|}
\hline
\multirow{2}{*}{Model} & Fine-tuning & \multicolumn{4}{|c|}{Analogy Tasks} & \multicolumn{2}{|c|}{Similarity Tasks} \\
\cline{3-8}
& Scheme & S\&S & Acoustic & S\&S-Acoustic & Average & Word & Acoustic 
\gdef\rownumber{\stepcounter{magicrownumbers}\arabic{magicrownumbers}} \\
\hline
Google W2V & - & 61.42\% & 0.9\% & 16.99\% & 26.44\% & \textbf{0.6893} & -0.3489 \\
In-domain W2V (556 dim.) & - & 63.6\% & 0.81\% & 14.54\% & 26.32\% & 0.6333 & -0.4717 \\
\hline
\multicolumn{8}{c}{Model Concatenation} \\
\hline
C2V-1 (F) + C2V-a (F) & - & 67.03\% & 25.43\% & 40.36\% & 44.27\% & 0.5102 & 0.7231 \\
C2V-1 (F) + Inter-Confusion (F) & - & 70.84\% & 35.25\% & 35.18\% & 47.09\% & 0.5609 & 0.6345 \\
C2V-1 (F) + Hybrid Intra-Inter (F) & - & 68.08\% & 11.39\% & 41.3\% & 40.26\% & 0.4142 & 0.5285 \\
\hline
\multicolumn{8}{c}{Fixed Contextual Subspace Joint Optimization} \\
\hline
C2V-1 (F) + C2V-a (L) & inter & 71.65\% & 20.54\% & 33.76\% & 41.98\% & 0.5676 & 0.4437 \\
C2V-1 (F) + C2V-a (L) & intra & 67.37\% & 28.64\% & 39.09\% & 45.03\% & 0.5211 & 0.6967 \\
C2V-1 (F) + C2V-a (L) & hybrid & 70.02\% & 25.84\% & 37.18\% & 44.35\% & 0.5384 & 0.6287 \\
C2V-1 (F) + C2V-c (L) & inter & 72.01\% & 35.25\% & 33.58\% & 46.95\% & 0.5266 & 0.5818 \\
C2V-1 (F) + C2V-c (L) & intra & 69.7\% & 39.32\% & 39.07\% & 49.36\% & 0.5156 & 0.7021 \\
C2V-1 (F) + C2V-c (L) & hybrid & \textbf{72.38\%} & 37.75\% & 37.95\% & 49.36\% & 0.5220 & 0.6674 \\
C2V-1 (F) + C2V-* (L) & inter & 71.36\% & 8.55\% & 33.21\% & 37.71\% & 0.5587 & 0.302 \\
C2V-1 (F) + C2V-* (L) & intra & 66.85\% & 13.33\% & 40.1\% & 40.09\% & 0.4996 & 0.5691 \\
C2V-1 (F) + C2V-* (L) & hybrid & 68.32\% & 11.61\% & 38.19\% & 39.37\% & 0.5254 & 0.4945 \\
\hline
\multicolumn{8}{c}{Unrestricted Joint Optimization} \\
\hline
C2V-1 (L) + C2V-a (L) & inter & 62.12\% & 46.42\% & 36.4\% & 48.31\% & 0.5513 & 0.7926 \\
C2V-1 (L) + C2V-a (L) & intra & 64.85\% & 40.55\% & 42.38\% & 49.26\% & 0.5033 & 0.7949 \\
C2V-1 (L) + C2V-a (L) & hybrid & 31.65\% & 61.91\% & 23.55\% & 39.04\% & $0.1067^{\ast}$ & \textbf{0.8309} \\
C2V-1 (L) + C2V-c (L) & inter & 64.98\% & 52.99\% & 34.79\% & 50.92\% & 0.5763 & 0.7725 \\
C2V-1 (L) + C2V-c (L) & intra & 65.88\% & 49.4\% & 41.51\% & \textbf{52.26\%} & 0.5379 & 0.7717 \\
C2V-1 (L) + C2V-c (L) & hybrid & 37.86\% & \textbf{67.21\%} & 25.96\% & 43.68\% & 0.2295 & 0.8294 \\
C2V-1 (L) + C2V-* (L) & inter & 65.54\% & 27.97\% & 36.87\% & 43.46\% & 0.5338 & 0.6953 \\
C2V-1 (L) + C2V-* (L) & intra & 64.42\% & 20.05\% & \textbf{42.56\%} & 42.34\% & 0.4920 & 0.6942 \\
C2V-1 (L) + C2V-* (L) & hybrid & 65.79\% & 22.63\% & 41.3\% & 43.24\% & 0.4967 & 0.6986 \\
\hline
\end{tabular}%
}
\end{center}
\captionsetup{justification=centering}
\caption{\textbf{Model concatenation and joint optimization results}\\
\small{\textbf{C2V-1: Top-Confusion, C2V-a: Intra-Confusion, C2V-c: Inter-Confusion, C2V-*: Hybrid Intra-Inter}\\
All the models are of 556 (300+256) dimensions.\\
Acronyms: (F):Fixed embedding, (L):Learn embedding during joint training, S\&S: Semantic \& Syntactic Analogy.\\
For the analogy tasks: the accuracies of baseline word2vec models are for top-1 evaluations, whereas of the other models are for top-2 evaluations (as discussed in Section~\ref{sec:evaluation_analogy_tasks}).
Detailed semantic analogy and syntactic analogy accuracies, the top-1 evaluations and top-2 evaluations for all the models are available under Appendix in Table~\ref{appendix:tab:joint_opt_results}.\\
For the similarity tasks: all the correlations (Spearman's) are statistically significant with $p<0.001$ except the ones with the asterisks.
Detailed $p-values$ for the correlations are presented under Appendix in Table~\ref{appendix:tab:joint_opt_results:similarity}.}}\label{tab:joint_opt_results}
\end{table*}

\subsection{Model Concatenation}
The first 4 rows of Table~\ref{tab:joint_opt_results} show the results with model concatenation.
We concatenate each of the proposed models (from Table~\ref{tab:results}) with the pre-trained top-confusion model (we use C2V-1 model instead of word2vec as hypothesized in Figure~\ref{fig:tuning_schemes}(b) because empirically C2V-1 model provided similar performance on Semantic\&Syntactic tasks and overall better average accuracy on analogy tasks compared to the baseline-in-domain W2V model).
Thus the resulting vector space is 556 dimensional (300 (pre-trained top-confusion model) + 256 (proposed models from Table~\ref{tab:results})).
In our case, we believe the dimension expansion of the vector space is insignificant in terms of performance considering the relatively low amount of training data compared to Google's word2vec model.
To be completely fair in judgment, we create a new baseline model with 556 dimensional embedding space for comparison.
To train the new baseline model, the 556 dimension embedding was initialized with 300 dimensional Google's word2vec embedding and the rest of the dimensions as zeros (null space).
Comparing the 556 dimension baseline from Table~\ref{tab:joint_opt_results} with the previous 300 dimensional baseline from Table~\ref{tab:pretraining_results}, the results are almost identical which confirms the dimension expansion is insignificant with respect to performance.

With model concatenation, we see slightly better results (average analogy accuracy) comparing with the pre-trained models from Table~\ref{tab:pretraining_results}, an absolute increase of up-to approximately 5\% among the best results.
The correlations with similarity tasks are similar and comparable with the earlier results with the pre-trained models.

\subsection{Joint Optimization}
\subsubsection{Fixed Contextual Subspace}
Rows 5-13 of Table~\ref{tab:joint_opt_results} display the results of joint optimization with concatenated, fixed top-confusion (C2V-1) embeddings and learn-able confusion2vec (C2V-a/c/*) embeddings.
As hypothesized with fixed subspace, the results indicate better accuracies for the Semantic\&Syntactic analogy task.
Thereby, the improvements also reflect on the overall average accuracy of the analogy tasks.
This also confirms the need for joint optimization which boosts the average accuracy up-to approximately 2\% absolute over the unoptimized concatenated model.

\subsubsection{Unrestricted Optimization}
The last 9 rows of Table~\ref{tab:joint_opt_results} display the results obtained by jointly optimizing the concatenated models without constraints.
Both the subspaces are fine tuned to convergence with various proposed training criteria.
We consistently observe improvements with the unrestricted optimization over the unoptimized model concatenations.
In terms of average accuracy we observe an increase in average accuracy by up-to 5\% (absolute) approximate over the unoptimized concatenated models.
Moreover, we obtain improvements over the Fixed Contextual Subspace joint-optimized models, up-to 2-3\% (absolute) in average accuracies.
The best overall model in terms of average accuracies is obtained by unrestricted joint optimization on the concatenated top-confusion and inter-confusion models by fine-tuning with the intra-confusion training scheme.

\subsection{Results Summary}
\Pg{Summary}
Firstly, comparing among the different training schemes (see Table~\ref{tab:results}), the inter-confusion training consistently gives the best Acoustic analogy accuracies, whereas the hybrid training scheme often gives the best Semantic\&Syntactic-Acoustic analogy accuracies.
As far as the Semantic\&Syntactic analogy task is concerned, the intra-confusion is often found to give preference to syntactic relations, while the inter-confusion boosts the semantic relations and the hybrid scheme balances both relations (see Appendix Table~\ref{appendix:tab:results}).
Next, pre-training/initializing the model gives drastic improvements in overall average accuracy of analogy tasks.
Concatenating the top-confusion model with the confusion2vec (C2V-a/c/*) model gives slightly better results.
More optimizations and fine-tuning over the concatenated model gives considerably the best results. 

\Pg{Highlights}
Overall, the best results are obtained with unrestricted joint optimization of top-confusion and inter-confusion model, i.e., fine-tuning using intra-confusion training mode.
In terms of average analogy accuracies the confusion2vec model (C2V-a/c/*) outperforms the baseline by up-to 26.06\%. 
The best performing confusion2vec model outperforms the word2vec model even on the Semantic\&Syntactic analogy tasks (by a relative 7.8\%).
Moreover, even the comparison between the top-2 evaluations of both the word2vec and confusion2vec (C2V-1/a/c/*) suggests very similar performance on Semantic\&Syntactic-analogy tasks (see Appendix Table~\ref{appendix:tab:joint_opt_results}).
This confirms and emphasizes that the confusion2vec (C2V-1/a/c/*) doesn't compromise on the information captured by word2vec but succeeds in augmenting the space with word confusions.
Another highlight observation is that modeling the word confusions boost the semantic and syntactic scores of the Semantic\&Syntactic analogy task (compared to word2vec), suggesting inherent information in word confusions which could be exploited for better semantic-syntactic word relation modeling.

\section{Vector Space Analysis}\label{sec:analysis}

In this section, we compare the vector space plots of the typical word2vec space and the proposed confusion2vec vector space for specifically chosen set of words.
We choose a subset of words representing three categories to reflect semantic relationships, syntactic relationships and acoustic relationships.
The vector space representations of the words are then subjected to dimension reduction using principle component analysis (PCA) to obtain 2D vectors which are used for plotting.
\subsection{Semantic Relationships}
\begin{figure*}[t]
\includegraphics[width=\linewidth]{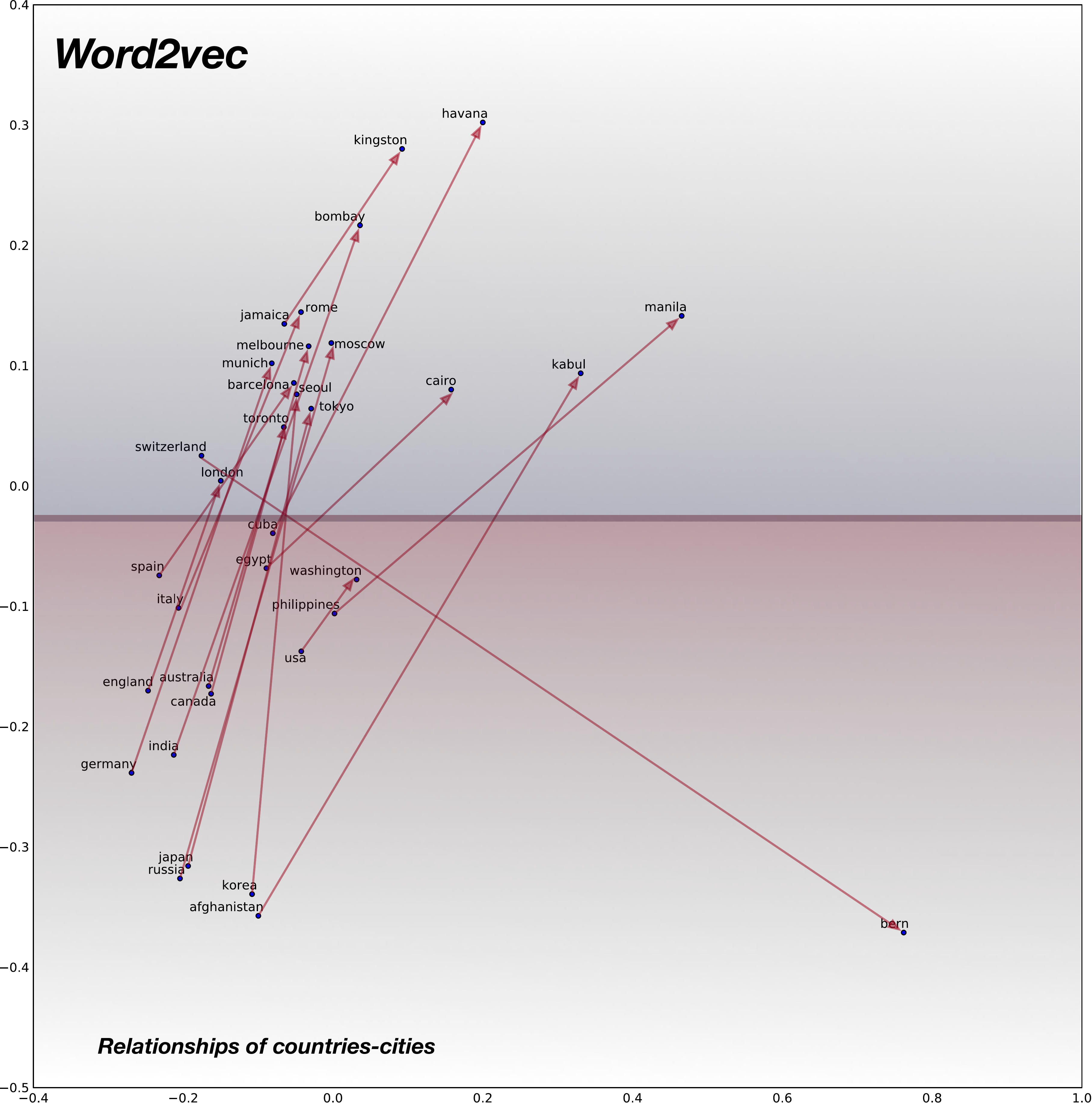}
\captionsetup{justification=centering}
\caption{\textbf{2D plot after PCA of word vector representation on baseline pre-trained word2vec\\
Demonstration of Semantic Relationship on Randomly chosen pairs of Countries and Cities}\\
\small{Country-City vectors are almost parallel/similar. Countries are clustered together on the bottom half (highlighted with red hue) and the cities on upper half (highlighted with blue hue).}}\label{fig:country_city_currency_w2v}
\end{figure*}

\begin{figure*}[t]
\includegraphics[width=\linewidth]{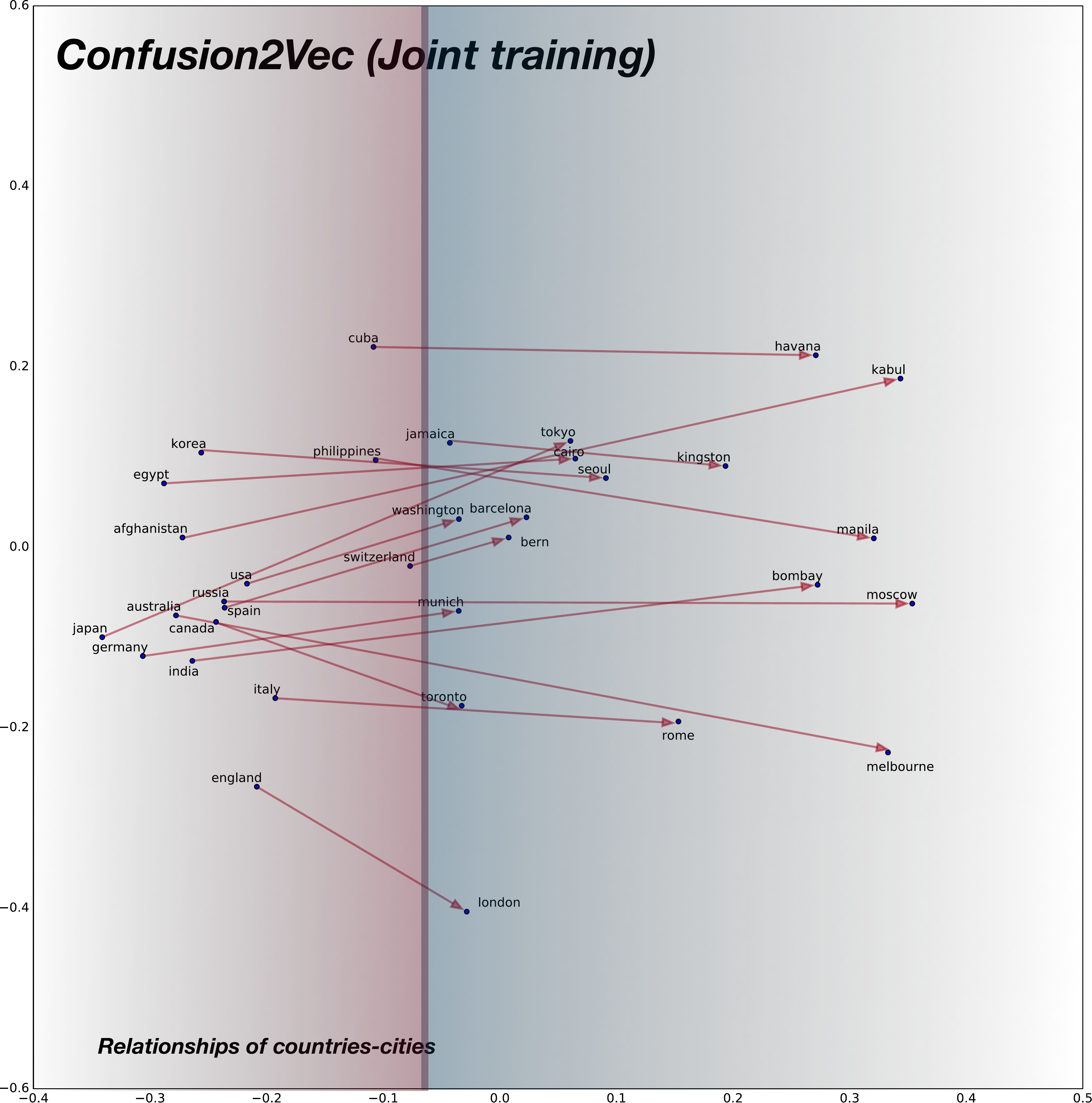}
\captionsetup{justification=centering}
\caption{\textbf{2D plot after PCA of word vector representation on jointly optimized pre-trained C2V-1 + C2V-a models\\
Demonstration of Semantic Relationship on Randomly chosen pairs of Countries and Cities}\\
\small{Observe the semantic relationships are preserved as in the case of word2vec model: Country-City vectors are almost parallel/similar. Countries are clustered together on the left half (highlighted with red hue) and the cities on right half (highlighted with blue hue).}}\label{fig:country_city_currency_C2V}
\end{figure*}

For analyzing the semantic relationships, we compile random word pairs (constrained by the availability of these in our training data) representing Country-Cities relationships.
The 2D plot for baseline pre-trained word2vec model is shown in Figure~\ref{fig:country_city_currency_w2v} and for the proposed confusion2vec model, specifically for the randomly selected, jointly-optimized top-confusion + intra-confusion model (corresponding to row 6 in Table~\ref{tab:joint_opt_results}) is displayed in Figure~\ref{fig:country_city_currency_C2V}.
The following observations can be made comparing the two PCA plots:
\begin{itemize}
\item Examining the baseline word2vec model, we find the Cities are clustered over the upper half of the plot (highlighted with blue hue in Figure~\ref{fig:country_city_currency_w2v}) and Countries are clustered together at the bottom half (highlighted with red hue in Figure~\ref{fig:country_city_currency_w2v}).
\item Similar trends are observed with the proposed confusion2vec model, where the cities are clustered together over the right half of the plot (highlighted with blue hue in Figure~\ref{fig:country_city_currency_C2V}) and the countries are grouped together towards the left half (highlighted with red hue in Figure~\ref{fig:country_city_currency_C2V}).
\item In the Word2Vec space, the vectors of Country-City word pairs are roughly parallel, pointing north-east (i.e., vectors are approximately similar). 
\item Similar to the word2vec space, with the Confusion2Vec, we observe the vectors of Country-City word pairs are fairly parallel and point to the east (i.e., vectors are highly similar).
\end{itemize}
The four observations indicate that the Confusion2Vec preserves the Semantic relationships between the words (similar to the Word2Vec space).

\subsection{Syntactic Relationships}
\begin{figure*}[t]
\includegraphics[width=\linewidth]{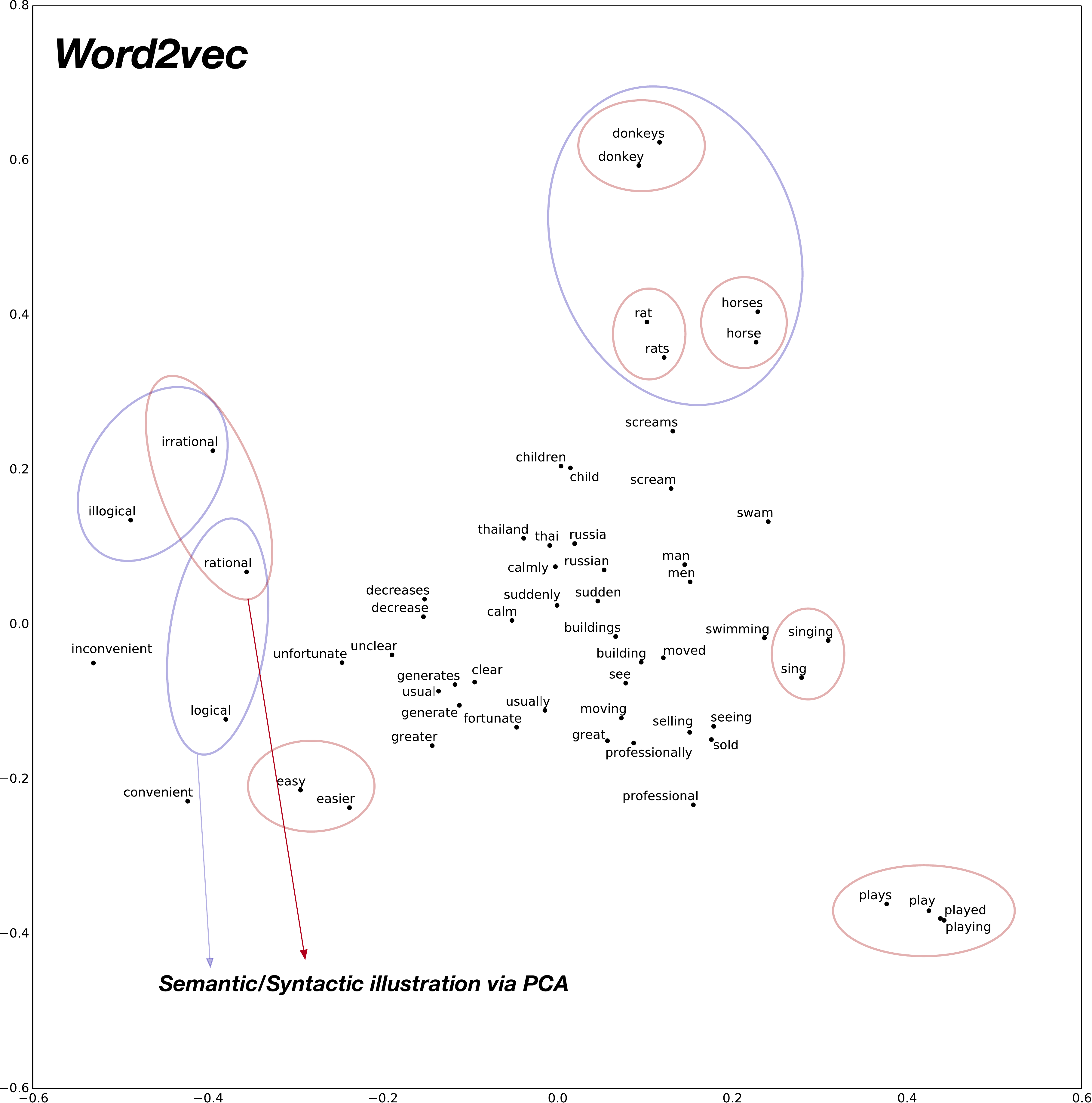}
\captionsetup{justification=centering}
\caption{\textbf{2D plot after PCA of word vector representation on baseline pre-trained word2vec\\
Demonstration of Syntactic Relationship on Randomly chosen 30 pairs of Adjective-Adverb, Opposites, Comparative, Superlative, Present-Participle, Past-tense, Plurals}\\
\small{Observe the clustering of syntactically related words (Ex: highlighted with red ellipses). Few semantically related words are highlighted with blue ellipses (Ex: animals)}}\label{fig:syntactic_w2v}
\end{figure*}

\begin{figure*}[t]
\includegraphics[width=\linewidth]{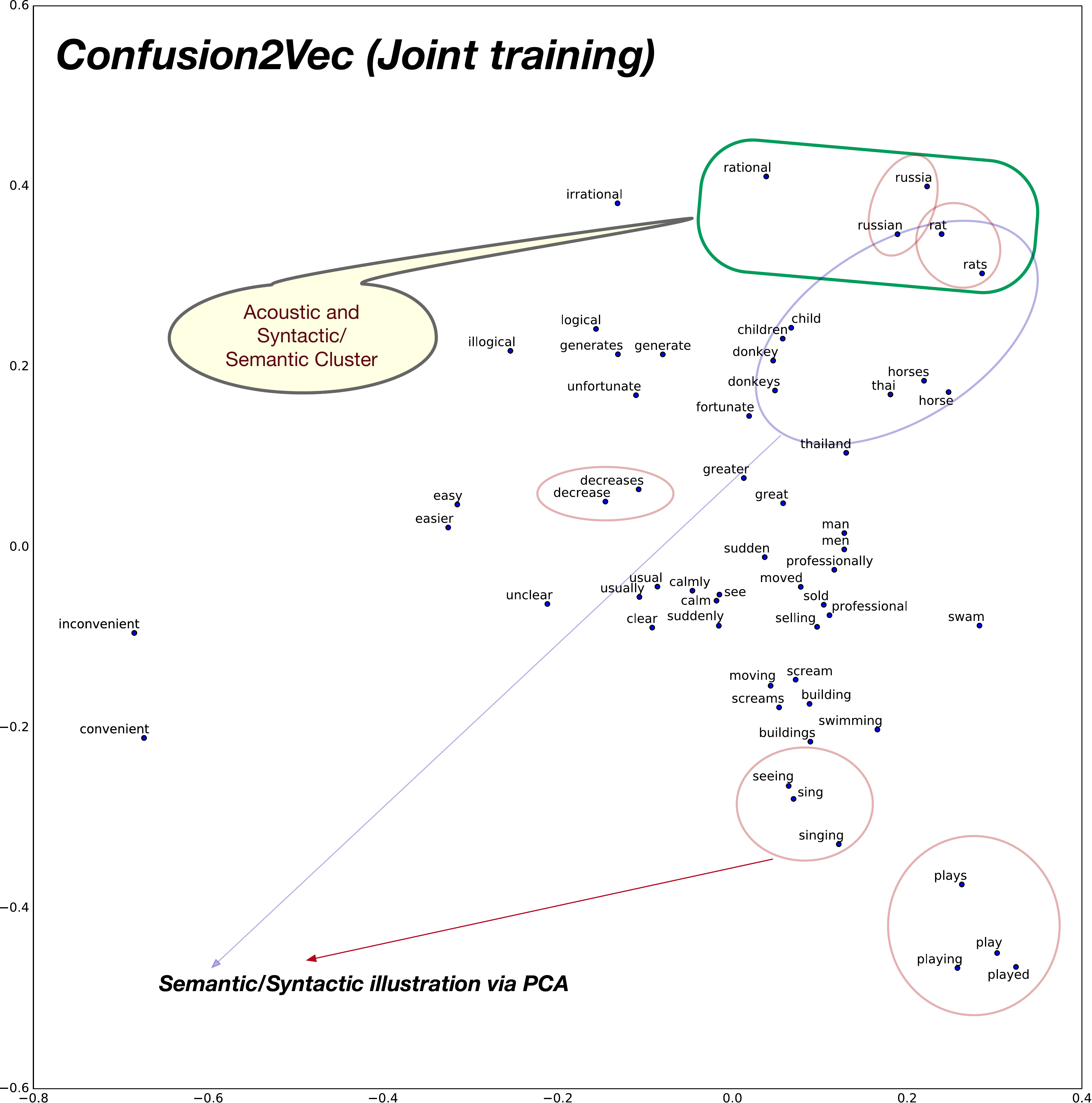}
\captionsetup{justification=centering}
\caption{\textbf{2D plot after PCA of word vector representation on jointly optimized pre-trained C2V-1 + C2V-a models\\
Demonstration of Syntactic Relationship on Randomly chosen 30 pairs of Adjective-Adverb, Opposites, Comparative, Superlative, Present-Participle, Past-tense, Plurals}\\
\small{Syntactic clustering is preserved by Confusion2Vec similar to Word2Vec. Red ellipses highlight few examples of syntactically related words. Similar to Word2Vec, semantically related words (Ex: animals), highlighted with blue ellipses, are also clustered together. Additionally Confusion2Vec clusters acoustically similar words together (indicated with green ellipse).}}\label{fig:syntactic_C2V}
\end{figure*}

To analyze the syntactic relationships, we create 30 pairs of words composed of Adjective-Adverb, Opposites, Comparative, Superlative, Present-Participle, Past-tense, Plurals.
The PCA 2D plots for baseline pre-trained word2vec model and the proposed confusion2vec model are illustrated in Figure~\ref{fig:syntactic_w2v} and Figure~\ref{fig:syntactic_C2V} respectively.
The following inferences can be made from the two plots:
\begin{itemize}
\item Inspecting the baseline word2vec model, we observe that the word pairs depicting syntactic relations occur often close-by (highlighted with red ellipses in Figure~\ref{fig:syntactic_w2v}). 
\item Few semantic relations are also apparent and are highlighted with blue ellipses in Figure~\ref{fig:syntactic_w2v}. For example, animals are clustered together.
\item Similarly, with the Confusion2Vec model, we observe syntactic clusters of words highlighted with red ellipses in Figure~\ref{fig:syntactic_C2V}.
\item Semantic relations apparent in the case of word2vec is also evident with the Confusion2Vec, which are highlighted with blue ellipses in Figure~\ref{fig:syntactic_C2V}.
\item Additionally with the Confusion2Vec model, we find clusters of acoustically similar words (with similar phonetic transcriptions). These are highlighted using a green ellipse in Figure~\ref{fig:syntactic_C2V}.
\end{itemize}
The above findings confirm that the confusion2vec models preserve the syntactic relationships similar to word2vec models, supporting our hypothesis.

\subsection{Acoustic Relationships}
\begin{figure*}[t]
\includegraphics[width=\linewidth]{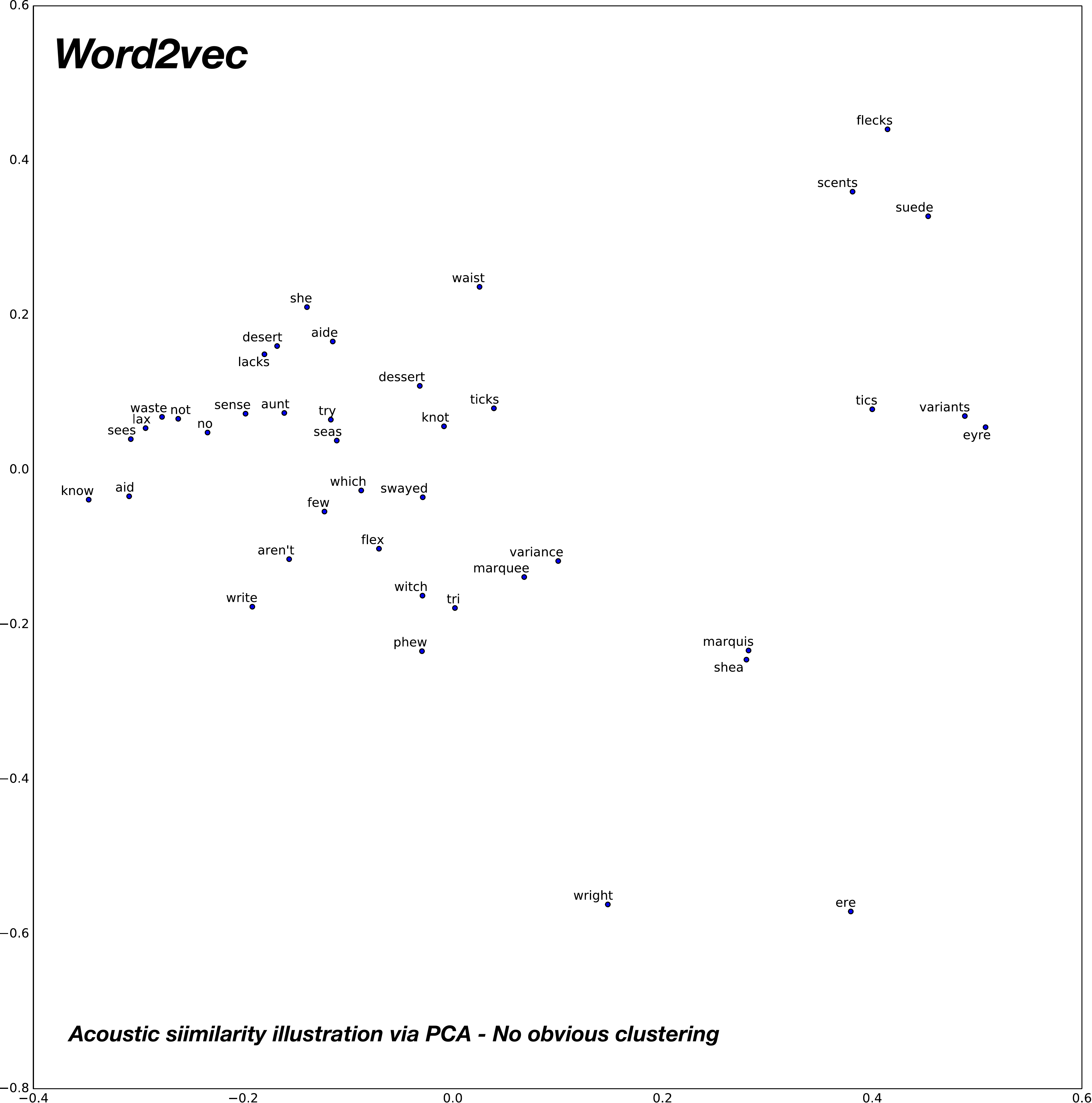}
\captionsetup{justification=centering}
\caption{\textbf{2D plot after PCA of word vector representation on baseline pre-trained word2vec\\
Demonstration of Vector Relationship on Randomly chosen 20 pairs of Acoustically Similar Sounding Words}\\
\small{No apparent relations between acoustically similar words are evident.}}\label{fig:homophones_w2v}
\end{figure*}

\begin{figure*}[t]
\includegraphics[width=\linewidth]{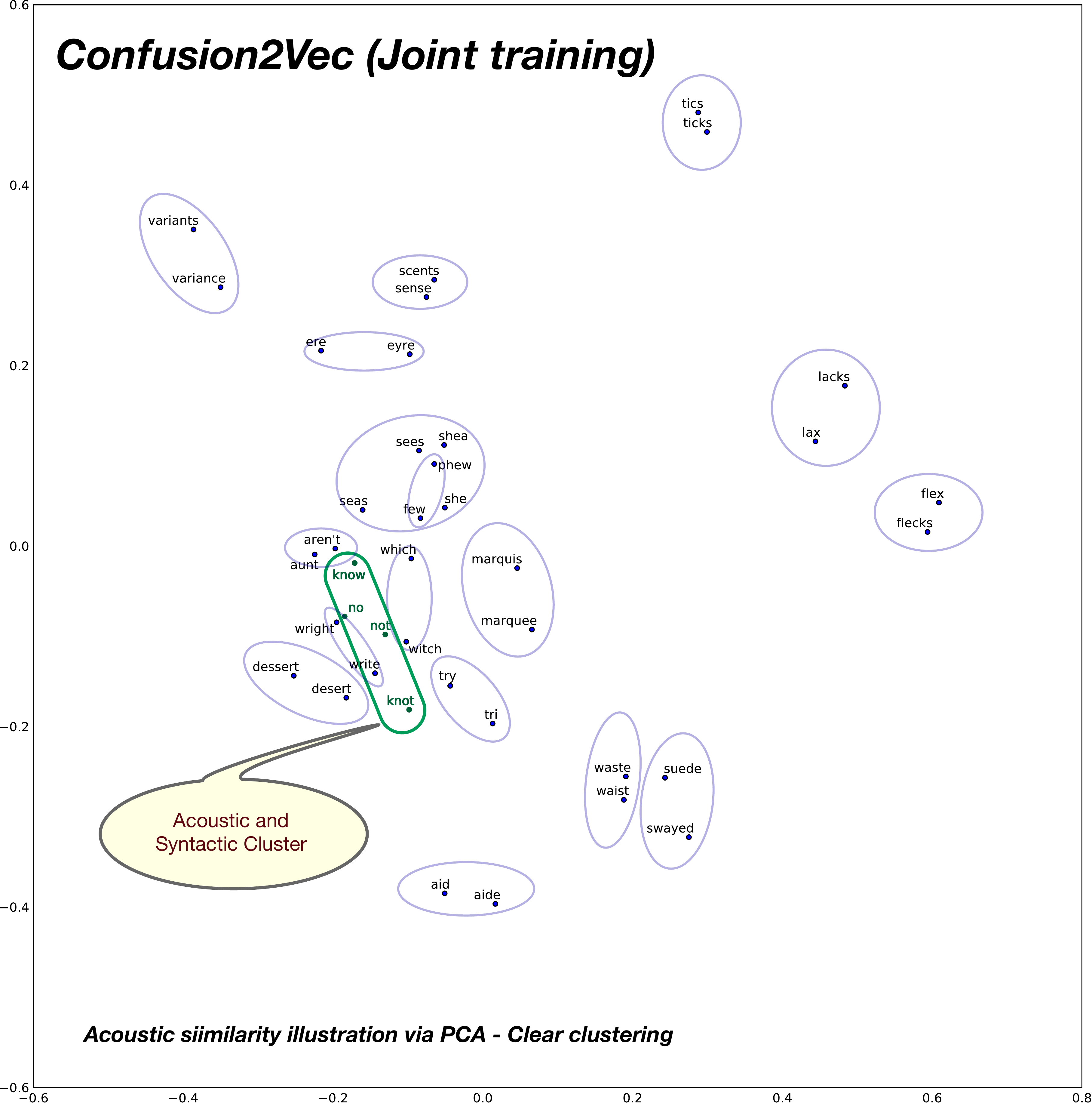}
\captionsetup{justification=centering}
\caption{\textbf{2D plot after PCA of word vector representation on jointly optimized pre-trained C2V-1 + C2V-a models\\
Demonstration of Vector Relationship on Randomly chosen 20 pairs of Acoustically Similar Sounding Words}\\
\small{Confusion2Vec clusters acoustically similar words together (highlighted with blue ellipses). Additionally, inter-relations between syntactically related words and acoustically related words are also evident (highlighted with green ellipse).}}\label{fig:homophones_C2V}
\end{figure*}

In order to analyze the relationships of similarly sounding words in the word vector spaces under consideration, we compose 20 pairs of acoustically similar sounding words, with similar phonetic transcriptions.
The 2D plot obtained after PCA for the baseline word2vec model is shown in Figure~\ref{fig:homophones_w2v} and the proposed confusion2vec model is shown in Figure~\ref{fig:homophones_C2V}.
We make the following observations from the two figures:
\begin{itemize}
\item Observing the baseline Word2vec model, no apparent trends are found between the acoustically similar words. For example, there is no trivial relationships apparent from the plot in Figure~\ref{fig:homophones_w2v} between the word ``no'' and ``know'', ``try'' and ``tri'' etc.
\item However, inspecting the proposed confusion2vec model, there is an obvious trend apparent, the acoustically similar words are grouped together in pairs and occur roughly in similar distances. The word pairs are highlighted with blue ellipses in Figure~\ref{fig:homophones_C2V}.
\item Additionally, in the Figure~\ref{fig:homophones_C2V}, as highlighted in green ellipse, we observe the 4 words ``no'', ``not'', ``knot'' and ``know'' occur in close proximity. The word pair ``no'' and ``not'' portray Semantic/Syntactic relation whereas the pairs ``knot'' \& ``not'' and ``no'' \& ``know'' are acoustically related.
\end{itemize}
The above findings suggest that the word2vec baseline model fails to capture any acoustic relationships whereas the proposed confusion2vec successfully models the confusions present in the lattices, in our specific case the acoustic confusions from the ASR lattices.

\section{Discussion}\label{sec:discussion}

\begin{table*}[t]
\begin{center}
\resizebox{\linewidth}{!}{
\setlength{\tabcolsep}{15pt}
\begin{tabular}{|c|c|c|c|c|}
\hline
Example & Ground-truth & ASR output & W2V Similarity & C2V Similarity \\
\hline
\newtag{1.1}{lab:1.1} & ``yes right answer'' & ``yes [right/write] answer'' & 0.96190 & 0.96218 \\
\newtag{1.2}{lab:1.2} & ``yes right answer'' & ``yes write answer'' & 0.93122 & 0.93194 \\
\newtag{1.3}{lab:1.3} & ``yes write answer'' & ``yes [right/write] answer'' & 0.99538 & 0.99548 \\
\newtag{1.4}{lab:1.4} & ``yes rite answer'' & ``yes [right/write] answer'' & 0.84216 & 0.88206 \\
\newtag{1.5}{lab:1.5} & ``yes rite answer'' & ``yes right answer'' & 0.86003 & 0.87085 \\
\newtag{1.6}{lab:1.6} & ``yes rite answer'' & ``yes write answer'' & 0.82073 & 0.87034 \\
\hline
\newtag{2.1}{lab:2.1} & ``she likes sea'' & ``[she/shea] likes [see/sea]'' & 0.91086 & 0.92130 \\
\newtag{2.2}{lab:2.2} & ``she likes sea'' & ``shea likes see'' & 0.73295 & 0.77137 \\
\newtag{2.3}{lab:2.3} & ``shea likes see'' & ``[she/shea] likes [see/sea]'' & 0.94807 & 0.95787 \\
\newtag{2.4}{lab:2.4} & ``shea likes see'' & ``[she/shea] likes [see/rocket]'' & 0.93560 & 0.93080 \\
\newtag{2.5}{lab:2.5} & ``she likes sea'' & ``[she/shea] likes [see/rocket]'' & 0.85853 & 0.85757 \\
\hline
\end{tabular}}
\captionsetup{justification=centering}
\caption{\textbf{Cosine Similarity between the ASR Ground-truth and ASR output in application to ASR error correction for baseline pre-trained word2vec and the proposed confusion2vec: jointly optimized intra-confusion + top-confusion models}\\
\small{Example 1.1-1.6 inherits structure as in Figure~\ref{fig:toy}(a), i.e., ``yes [right/write] answer'' assigns weight of 1.0 to yes and answer, 0.75 to right, 0.25 to write. Similarly Example 2.1-2.5 inherits structure as in Figure~\ref{fig:toy}(b)}}\label{tab:hyp_tests}
\end{center}
\end{table*}


\begin{figure}[b]
\centering
\includegraphics[width=\linewidth]{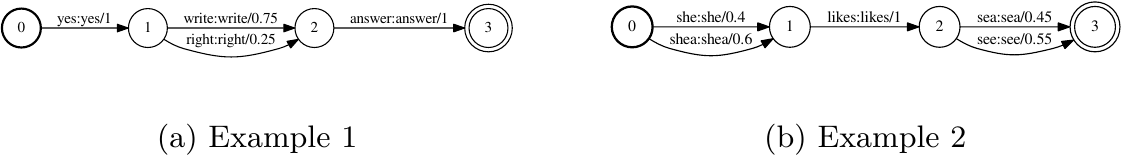}
\captionsetup{justification=centering}
\caption{\textbf{Confusion Network Examples}}\label{fig:toy}
\end{figure}

\begin{figure}[b]
\centering
\includegraphics[width=0.8\linewidth]{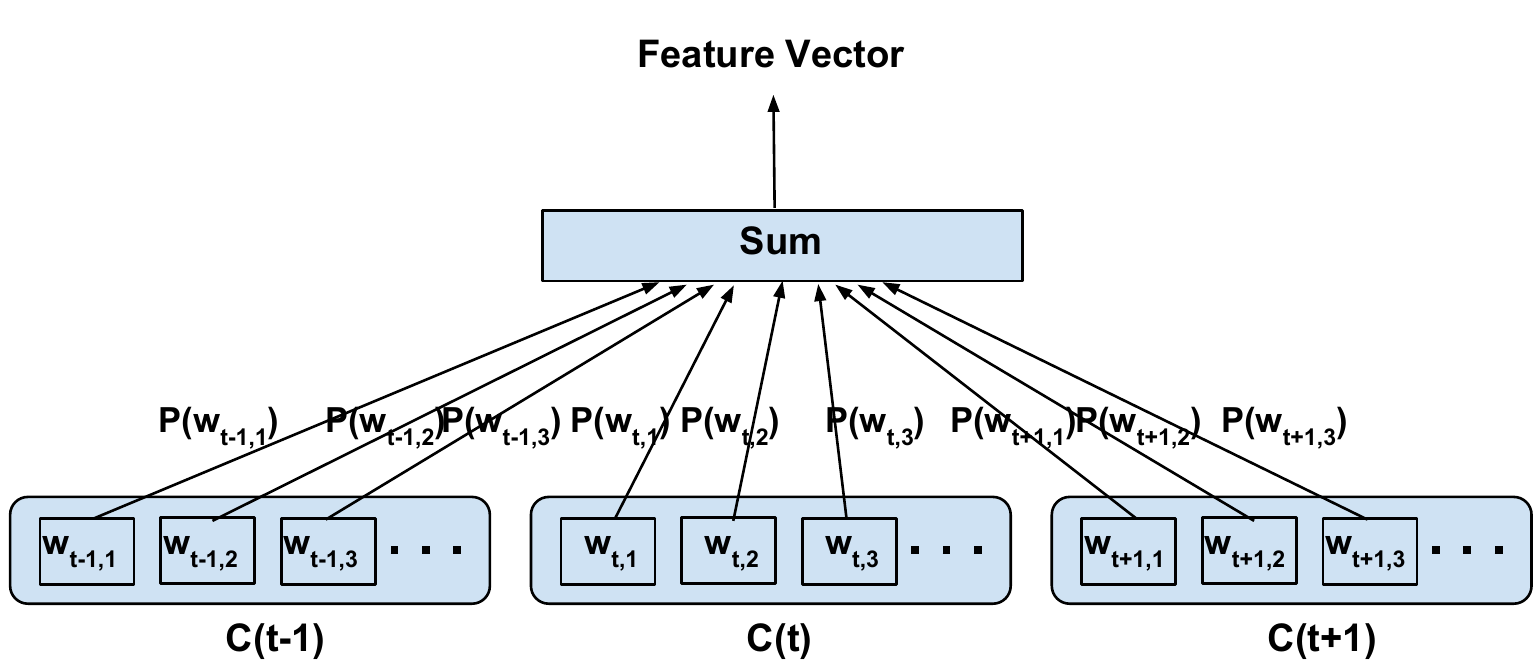}
\captionsetup{justification=centering}
\caption{\textbf{Computation of lattice feature vector.}}\label{fig:sol_arch}
\end{figure}

In this section, we demonstrate why the proposed embedding space is superior for modeling word lattices with the support of toy examples.
Lets consider a simple task of ASR error correction.
As shown by \cite{allauzen2007error,ogata2005speech,shivakumar2018learning}, often, the information needed to correct the errors are embedded in the lattices.
The toy examples in Figure~\ref{fig:toy}(a) \&~\ref{fig:toy}(b) depict the real scenarios encountered in ASR.
The lattice feature representation is a weighted vector sum of all words in the confusion and its context present in the lattice (see Figure~\ref{fig:sol_arch}).
We compare the proposed confusion2vec embeddings with the popular word2vec using cosine similarity as the evaluation measure.
Table~\ref{tab:hyp_tests} lists the evaluation for the following cases: (i)  ASR output is correct, (ii) ASR output is wrong and the correct candidate is present in the lattice, (iii) ASR output is wrong and the correct candidate is absent from the lattice, and (iv) ASR output is wrong and with no lattice available.
The following observations are drawn from the results:
\begin{enumerate}
\item Confusion2vec shows higher similarity with the correct answers when the ASR output is correct (see Table~\ref{tab:hyp_tests} example~\ref{lab:1.1},~\ref{lab:2.1}).
\item Confusion2vec exhibits higher similarity with the correct answers when the ASR output is wrong - meaning the representation is closer to the correct candidate and therefore more likely to correct the errors (see Table~\ref{tab:hyp_tests} example~\ref{lab:1.2},~\ref{lab:2.2},~\ref{lab:1.3},~\ref{lab:2.3}).
\item Confusion2vec yields high similarity even when the correct word candidate is not present in the lattice - meaning confusion2vec leverages inherent word representation knowledge to aid re-introduction of pruned or unseen words during error correction (see Table~\ref{tab:hyp_tests} example~\ref{lab:1.4},~\ref{lab:1.5},~\ref{lab:1.6}).
\item The confusion2vec shows low similarity in the case of fake lattices with highly unlikely word alternatives (see Table~\ref{tab:hyp_tests} example~\ref{lab:2.4},~\ref{lab:2.5}).
\end{enumerate}
All the above observations are supportive of the proposed confusion2vec word representation and is in line with the expectations for the task of ASR error correction.

\section{Potential Applications}\label{sec:apps}

In addition to the above discussed ASR error correction task, other potential application include:
\vspace{2mm}

\heading{Machine Translation}
In Machine Translation, word lattices are used to provide multiple sources for generating a single translation \citep{schroeder2009word,dyer2010formal}.
Word lattices derived from reordered hypotheses \citep{costa2007analysis,niehues2009pos,hardmeier2010fbk}, morphological transformations \citep{dyer2007noisier,hardmeier2010fbk}, word segmentations \citep{dyer2009using}, paraphrases \citep{onishi2011paraphrase} are used to introduce ambiguity and alternatives for training machine translation systems \citep{wuebker2012phrase,dyer2008generalizing,dyer2010formal}. Source language alternatives can also be exploited by introducing ambiguity derived from the combination of multiple machine translation systems \citep{matusov2006computing,rosti2007improved,rosti2007combining}. 
\emph{In the case of Machine Translation, the word-confusion subspace is associated with morphological transformations, word segmentations, paraphrases, part-of-speech information, etc., or a combination of them.}
Although the word-confusion subspace is not orthogonal, the explicit modeling of such ambiguity relationships is beneficial.

\heading{NLP}
Other NLP based applications like paraphrase generation \citep{quirk2004monolingual}, word segmentation \citep{kruengkrai2009error}, part-of-speech tagging \citep{kruengkrai2009error} also operate on lattices.
As discussed in section~\ref{sec:machine}, confusion2vec can exploit the ambiguity present in the lattices for betterment of the tasks.

\heading{ASR}
In ASR systems, word lattices and confusion networks are often re-scored using various algorithms to improve their performances by exploiting ambiguity \citep{sundermeyer2014lattice,mangu2000finding,xiong2016achieving,liu2014efficient}. 
\emph{In the case of ASR, the word-confusion subspace is associated with acoustic similarity of words which is often orthogonal to the semantic-syntactic subspace as discussed in section~\ref{sec:human}}.
~\ref{ex:1},~\ref{ex:2} and~\ref{ex:3} are prime cases supporting the need for jointly modeling acoustic word confusions and semantic-syntactic subspace.

\heading{Spoken Language Understanding}
Similarly, as in the case of ASR, Confusion2Vec could exploit the inherent acoustic word-confusion information for keyword spotting \citep{mangu2000finding}, confidence score estimation \citep{mangu2000finding,seigel2011combining,kemp1997estimating,jiang2005confidence}, domain adaptation \citep{shivakumar2018learning}, lattice compression \citep{mangu2000finding}, spoken content retrieval \citep{chelba2008retrieval,hori2007open}, system combinations \citep{mangu2000finding,hoffmeister2007cross} and other spoken language understanding tasks \citep{hakkani2006beyond,tur2002improving,marin2012using} which operate on lattices.

\heading{Speech Translation}
In speech translation systems, incorporating the word lattices and confusion networks (instead of the single top hypothesis) is beneficial in better integrating speech recognition system to the machine translation systems \citep{bertoldi2007speech,mathias2006statistical,matusov2005integration,schultz2004using}.
Similarly, exploiting uncertainty information between the ``ASR - Machine Translation - Speech synthesis'' systems for Speech-to-speech translation is useful \citep{lavie1997janus,wahlster2013verbmobil}.
Since speech translation involves combination of ASR and the Machine Translation systems, the word-confusion subspace is associated with a combination of acoustic word-similarity (for ASR) and morphological-segmentation-paraphrases ambiguities (for Machine Translation).
\begin{equation}
``\text{See son winter is here}'' \longrightarrow ``\text{voir fils hiver est ici}'' \tag{Example 4} \label{ex:4}
\end{equation}
\begin{equation}
``\text{Season winter is here}'' \longrightarrow ``\text{saison hiver est ici}'' \tag{Example 5} \label{ex:5}
\end{equation}
~\ref{ex:4} and~\ref{ex:5} demonstrate a case of speech translation of identically sounding English phrases to French.
Words ``See son'' and ``Season'' demonstrate ambiguity in terms of word segmentation.
Whereas the phrases ``See son'' and ``Season'' also exhibit ambiguity in terms of acoustic similarity.
By modeling both word-segmentation and acoustic-confusion through word vector representations, the confusion2vec can provide crucial information that the french words ``voir'' and ``saison'' are confusable under speech translation framework.

\heading{Optical Character Recognition}
In optical character recognition (OCR) systems, the confusion axis is related to pictorial structures of the characters/words. 
For example, say the characters ``{\fontfamily{pag}\selectfont a}'' and ``{\fontfamily{pag}\selectfont o}'' are easily confusable thus leading to similar character vectors in the embedding space.
In the case of word level confusions leading to words ``{\fontfamily{pag}\selectfont ward}'' and ``{\fontfamily{pag}\selectfont word}'' being similar with confusion2vec (word2vec would have the words ``word'' and ``ward'' fairly dissimilar).
Having this crucial optical confusion information is useful during OCR decoding on sequence of words when used in conjunction with the linguistic contextual information.

\heading{Image/Video Scene Summarization}
The task of scene summarization involves generating descriptive text summarizing the content in one or more images.
Intuitively, the task would benefit from linguistic contextual knowledge during the text generation.
However, with the confusion2vec, one can model and expect to capture two additional information streams (i) pictorial confusion of image/object recognizer, and (ii) pictorial context, i.e., modeling objects occurring together (e.g. we can expect oven to often appear nearby a stove or other kitchen appliances).
The additional streams of valuable information embedded in the lattices can contribute for better decoding.
In other words, for example, word2vec can exhibit high dissimilarity between the words ``lifebuoy'' and ``donuts'', however the confusion2vec can capture their pictorial similarity in a better word space representation and thus aiding in their end application of scene summarization.

\section{Conclusion}\label{sec:conclusion}
In this work, we proposed a new word vector representation motivated from human speech \& perception and aspects of machine learning for incorporating word confusions from lattice like structures. 
The proposed confusion2vec model is meant to capture additional word-confusion information and improve upon the popular word2vec models without compromising the inherent information captured by the word2vec models.
Although the word confusions could be domain/task specific, we present a case study on ASR lattices where the confusions are based on acoustic similarity of words.
Specifically, with respect to ASR related applications, the aim is to capture the contextual statistics, as with word2vec, and additionally also capture the acoustic word confusion statistics.
Several training configurations are proposed for confusion2vec model, each utilizating different degrees of acoustic confusability versus contextual information, present in the noisy (confusion network) ASR output, for modeling the word vector space.
Further, techniques like pre-training/initializations, model concatenation and joint optimization are proposed and evaluated for the confusion2vec models.
Appropriate evaluation schemes are formulated for the domain specific application.
The evaluation schemes are inspired from the popular analogy based question test set and word similarity tasks.
A new analogy task and word similarity tasks are designed for the acoustic confusion/similarity scenario. 
A detailed tabulation of results are presented for the confusion2vec model and compared to the baseline word2vec models.

The results show that the confusion2vec can augment additional task-specific word confusion information without compromising on the semantic and syntactic relationships captured by the word2vec models.
Next, detailed analysis is conducted on the confusion2vec vector space through PCA reduced 2-dimensional plots for three independent word relations: (i) Semantic relations, (ii) Syntactic relations, and (iii) Acoustic relations.
The analysis further supports our aforementioned experimental inferences.
Few toy examples are presented towards the task of ASR error correction to support the adequacy of the Confusion2vec over the word2vec word representations.
The study validates through various hypotheses and test results, the potential benefits of the confusion2vec model.

\section{Future Work}\label{sec:future_work}
In future, we plan to work on improving the confusion2vec model by incorporating the sub-word and phonemic transcription of words during training.
Sub-words and character transcription information is shown to improve the word vector representation \citep{bojanowski2017enriching,chen2015joint}.
We believe the sub-words and phoneme transcriptions of words are even more relevant to confusion2vec.
In addition to the improvements expected towards the semantic and syntactic representations (word2vec), since the sub-words and phoneme transcriptions of acoustically similar words are similar, it should help in modeling the confusion2vec to a much greater extent.

Apart from concentrating on improving the confusion2vec model, this work opens new possible opportunities in incorporating the confusion2vec embeddings to a whole range of full-fledged applications such as ASR error correction, speech translation tasks, machine translation, discriminative language models, optical character recognition, image/video scene summarization etc.

\section*{Acknowledgments}
The U.S. Army Medical Research Acquisition Activity, 820 Chandler Street, Fort Detrick MD 21702- 5014 is the awarding and administering acquisition office. This work was supported by the Office of the Assistant Secretary of Defense for Health Affairs through the Psychological Health and Traumatic Brain Injury Research Program under Award No. W81XWH-15-1-0632. 

\clearpage

\clearpage

\clearpage

\clearpage

\clearpage

\clearpage

\clearpage
\newpage
\balance
\bibliographystyle{plainnat}
\bibliography{refs}

\begin{thebibliography}{}

\bibitem[Abadi et~al., 2016]{abadi2016tensorflow}
Abadi, M., Barham, P., Chen, J., Chen, Z., Davis, A., Dean, J., Devin, M.,
  Ghemawat, S., Irving, G., Isard, M., et~al. (2016).
\newblock Tensorflow: a system for large-scale machine learning.
\newblock In {\em OSDI}, volume~16, pages 265--283.

\bibitem[Allauzen, 2007]{allauzen2007error}
Allauzen, A. (2007).
\newblock Error detection in confusion network.
\newblock In {\em Eighth Annual Conference of the International Speech
  Communication Association}.

\bibitem[Bengio and Heigold, 2014]{bengio2014}
Bengio, S. and Heigold, G. (2014).
\newblock Word embeddings for speech recognition.
\newblock In {\em Proceedings of the 15th Conference of the International
  Speech Communication Association, {Interspeech}}.

\bibitem[Bengio et~al., 2003]{bengio2003neural}
Bengio, Y., Ducharme, R., Vincent, P., and Jauvin, C. (2003).
\newblock A neural probabilistic language model.
\newblock {\em Journal of machine learning research}, 3(Feb):1137--1155.

\bibitem[Bertoldi et~al., 2007]{bertoldi2007speech}
Bertoldi, N., Zens, R., and Federico, M. (2007).
\newblock Speech translation by confusion network decoding.
\newblock In {\em Acoustics, Speech and Signal Processing, 2007. ICASSP 2007.
  IEEE International Conference on}, volume~4, pages IV--1297. IEEE.

\bibitem[Blei et~al., 2003]{blei2003latent}
Blei, D.~M., Ng, A.~Y., and Jordan, M.~I. (2003).
\newblock Latent dirichlet allocation.
\newblock {\em Journal of machine Learning research}, 3(Jan):993--1022.

\bibitem[Bojanowski et~al., 2017]{bojanowski2017enriching}
Bojanowski, P., Grave, E., Joulin, A., and Mikolov, T. (2017).
\newblock Enriching word vectors with subword information.
\newblock {\em Transactions of the Association for Computational Linguistics},
  5:135--146.

\bibitem[Botha and Blunsom, 2014]{botha2014compositional}
Botha, J. and Blunsom, P. (2014).
\newblock Compositional morphology for word representations and language
  modelling.
\newblock In {\em International Conference on Machine Learning}, pages
  1899--1907.

\bibitem[Buckman and Neubig, 2018]{buckman2018neural}
Buckman, J. and Neubig, G. (2018).
\newblock Neural lattice language models.
\newblock {\em arXiv preprint arXiv:1803.05071}.

\bibitem[Celebi et~al., 2012]{celebi2012semi}
Celebi, A., Sak, H., Dikici, E., Sara{\c{c}}lar, M., Lehr, M., Prud'hommeaux,
  E., Xu, P., Glenn, N., Karakos, D., Khudanpur, S., et~al. (2012).
\newblock Semi-supervised discriminative language modeling for {T}urkish {ASR}.
\newblock In {\em Acoustics, Speech and Signal Processing (ICASSP), 2012 IEEE
  International Conference on}, pages 5025--5028. IEEE.

\bibitem[Chelba et~al., 2008]{chelba2008retrieval}
Chelba, C., Hazen, T.~J., and Saraclar, M. (2008).
\newblock Retrieval and browsing of spoken content.
\newblock {\em IEEE Signal Processing Magazine}, 25(3).

\bibitem[Chen et~al., 2015]{chen2015joint}
Chen, X., Xu, L., Liu, Z., Sun, M., and Luan, H.-B. (2015).
\newblock Joint learning of character and word embeddings.
\newblock In {\em IJCAI}, pages 1236--1242.

\bibitem[Chung et~al., 2016]{chung2016audio}
Chung, Y.-A., Wu, C.-C., Shen, C.-H., Lee, H.-Y., and Lee, L.-S. (2016).
\newblock Audio word2vec: Unsupervised learning of audio segment
  representations using sequence-to-sequence autoencoder.
\newblock {\em arXiv preprint arXiv:1603.00982}.

\bibitem[Cieri et~al., 2004]{cieri2004fisher}
Cieri, C., Miller, D., and Walker, K. (2004).
\newblock The fisher corpus: a resource for the next generations of
  speech-to-text.
\newblock In {\em LREC}, volume~4, pages 69--71.

\bibitem[Costa-juss{\`a} and Fonollosa, 2007]{costa2007analysis}
Costa-juss{\`a}, M.~R. and Fonollosa, J.~A. (2007).
\newblock Analysis of statistical and morphological classes to generate
  weighted reordering hypotheses on a statistical machine translation system.
\newblock In {\em Proceedings of the Second Workshop on Statistical Machine
  Translation}, pages 171--176. Association for Computational Linguistics.

\bibitem[Cotterell and Sch{\"u}tze, 2015]{cotterell2015morphological}
Cotterell, R. and Sch{\"u}tze, H. (2015).
\newblock Morphological word-embeddings.
\newblock In {\em Proceedings of the 2015 Conference of the North American
  Chapter of the Association for Computational Linguistics: Human Language
  Technologies}, pages 1287--1292.

\bibitem[Deerwester et~al., 1990]{deerwester1990indexing}
Deerwester, S., Dumais, S.~T., Furnas, G.~W., Landauer, T.~K., and Harshman, R.
  (1990).
\newblock Indexing by latent semantic analysis.
\newblock {\em Journal of the American society for information science},
  41(6):391.

\bibitem[Dikici et~al., 2012]{dikici2012performance}
Dikici, E., Celebi, A., and Sara{\c{c}}lar, M. (2012).
\newblock Performance comparison of training algorithms for semi-supervised
  discriminative language modeling.
\newblock In {\em Thirteenth Annual Conference of the International Speech
  Communication Association}.

\bibitem[Dyer, 2009]{dyer2009using}
Dyer, C. (2009).
\newblock Using a maximum entropy model to build segmentation lattices for mt.
\newblock In {\em Proceedings of Human Language Technologies: The 2009 Annual
  Conference of the North American Chapter of the Association for Computational
  Linguistics}, pages 406--414. Association for Computational Linguistics.

\bibitem[Dyer et~al., 2008]{dyer2008generalizing}
Dyer, C., Muresan, S., and Resnik, P. (2008).
\newblock Generalizing word lattice translation.
\newblock Technical report, MARYLAND UNIV COLLEGE PARK INST FOR ADVANCED
  COMPUTER STUDIES.

\bibitem[Dyer, 2007]{dyer2007noisier}
Dyer, C.~J. (2007).
\newblock The'noisier channel': translation from morphologically complex
  languages.
\newblock In {\em Proceedings of the Second Workshop on Statistical Machine
  Translation}, pages 207--211. Association for Computational Linguistics.

\bibitem[Dyer, 2010]{dyer2010formal}
Dyer, C.~J. (2010).
\newblock {\em A formal model of ambiguity and its applications in machine
  translation}.
\newblock University of Maryland, College Park.

\bibitem[Erhan et~al., 2010]{erhan2010does}
Erhan, D., Bengio, Y., Courville, A., Manzagol, P.-A., Vincent, P., and Bengio,
  S. (2010).
\newblock Why does unsupervised pre-training help deep learning?
\newblock {\em Journal of Machine Learning Research}, 11(Feb):625--660.

\bibitem[Faruqui and Dyer, 2014]{faruqui2014improving}
Faruqui, M. and Dyer, C. (2014).
\newblock Improving vector space word representations using multilingual
  correlation.
\newblock In {\em Proceedings of the 14th Conference of the European Chapter of
  the Association for Computational Linguistics}, pages 462--471.

\bibitem[Finkelstein et~al., 2001]{finkelstein2001placing}
Finkelstein, L., Gabrilovich, E., Matias, Y., Rivlin, E., Solan, Z., Wolfman,
  G., and Ruppin, E. (2001).
\newblock Placing search in context: The concept revisited.
\newblock In {\em Proceedings of the 10th international conference on World
  Wide Web}, pages 406--414. ACM.

\bibitem[Ghannay et~al., 2015a]{Ghannay:2015:CCW:2963447.2963456}
Ghannay, S., Est\`{e}ve, Y., Camelin, N., Dutrey, C., Santiago, F., and
  Adda-Decker, M. (2015a).
\newblock Combining continuous word representation and prosodic features for
  asr error prediction.
\newblock In {\em Proceedings of the Third International Conference on
  Statistical Language and Speech Processing - Volume 9449}, SLSP 2015, pages
  84--95, New York, NY, USA. Springer-Verlag New York, Inc.

\bibitem[Ghannay et~al., 2015b]{7362668}
Ghannay, S., Estève, Y., and Camelin, N. (2015b).
\newblock Word embeddings combination and neural networks for robustness in asr
  error detection.
\newblock In {\em 2015 23rd European Signal Processing Conference (EUSIPCO)},
  pages 1671--1675.

\bibitem[Ghannay et~al., 2016]{Ghannay_2016}
Ghannay, S., Estève, Y., Camelin, N., and deléglise, P. (2016).
\newblock Acoustic word embeddings for asr error detection.
\newblock In {\em Interspeech 2016}, pages 1330--1334.

\bibitem[Hakkani-T{\"u}r et~al., 2006]{hakkani2006beyond}
Hakkani-T{\"u}r, D., B{\'e}chet, F., Riccardi, G., and Tur, G. (2006).
\newblock Beyond asr 1-best: Using word confusion networks in spoken language
  understanding.
\newblock {\em Computer Speech \& Language}, 20(4):495--514.

\bibitem[Hardmeier et~al., 2010]{hardmeier2010fbk}
Hardmeier, C., Bisazza, A., and Federico, M. (2010).
\newblock Word lattices for morphological reduction and chunk-based reordering.
\newblock In {\em Proceedings of the Joint Fifth Workshop on Statistical
  Machine Translation and MetricsMATR}, pages 88--92. Association for
  Computational Linguistics.

\bibitem[He et~al., 2016]{he2016multi}
He, W., Wang, W., and Livescu, K. (2016).
\newblock Multi-view recurrent neural acoustic word embeddings.
\newblock {\em arXiv preprint arXiv:1611.04496}.

\bibitem[Hoffmeister et~al., 2007]{hoffmeister2007cross}
Hoffmeister, B., Hillard, D., Hahn, S., Schluter, R., Ostendor, M., and Ney, H.
  (2007).
\newblock Cross-site and intra-site asr system combination: Comparisons on
  lattice and 1-best methods.
\newblock In {\em Acoustics, Speech and Signal Processing, 2007. ICASSP 2007.
  IEEE International Conference on}, volume~4, pages IV--1145. IEEE.

\bibitem[Hofmann, 1999]{hofmann1999probabilistic}
Hofmann, T. (1999).
\newblock Probabilistic latent semantic analysis.
\newblock In {\em Proceedings of the Fifteenth conference on Uncertainty in
  artificial intelligence}, pages 289--296. Morgan Kaufmann Publishers Inc.

\bibitem[Hori et~al., 2007]{hori2007open}
Hori, T., Hetherington, I.~L., Hazen, T.~J., and Glass, J.~R. (2007).
\newblock Open-vocabulary spoken utterance retrieval using confusion networks.
\newblock In {\em Acoustics, Speech and Signal Processing, 2007. ICASSP 2007.
  IEEE International Conference on}, volume~4, pages IV--73. IEEE.

\bibitem[Huang et~al., 2012]{huang2012improving}
Huang, E.~H., Socher, R., Manning, C.~D., and Ng, A.~Y. (2012).
\newblock Improving word representations via global context and multiple word
  prototypes.
\newblock In {\em Proceedings of the 50th Annual Meeting of the Association for
  Computational Linguistics: Long Papers-Volume 1}, pages 873--882. Association
  for Computational Linguistics.

\bibitem[Jiang, 2005]{jiang2005confidence}
Jiang, H. (2005).
\newblock Confidence measures for speech recognition: A survey.
\newblock {\em Speech communication}, 45(4):455--470.

\bibitem[Joulin et~al., 2016]{joulin2016bag}
Joulin, A., Grave, E., Bojanowski, P., and Mikolov, T. (2016).
\newblock Bag of tricks for efficient text classification.
\newblock {\em arXiv preprint arXiv:1607.01759}.

\bibitem[Kamper et~al., 2016]{kamper2016deep}
Kamper, H., Wang, W., and Livescu, K. (2016).
\newblock Deep convolutional acoustic word embeddings using word-pair side
  information.
\newblock In {\em Acoustics, Speech and Signal Processing (ICASSP), 2016 IEEE
  International Conference on}, pages 4950--4954. IEEE.

\bibitem[Kemp and Schaaf, 1997]{kemp1997estimating}
Kemp, T. and Schaaf, T. (1997).
\newblock Estimating confidence using word lattices.
\newblock In {\em Fifth European Conference on Speech Communication and
  Technology}.

\bibitem[Kim, 2014]{kim2014convolutional}
Kim, Y. (2014).
\newblock Convolutional neural networks for sentence classification.
\newblock {\em arXiv preprint arXiv:1408.5882}.

\bibitem[Kruengkrai et~al., 2009]{kruengkrai2009error}
Kruengkrai, C., Uchimoto, K., Kazama, J., Wang, Y., Torisawa, K., and Isahara,
  H. (2009).
\newblock An error-driven word-character hybrid model for joint chinese word
  segmentation and pos tagging.
\newblock In {\em Proceedings of the Joint Conference of the 47th Annual
  Meeting of the ACL and the 4th International Joint Conference on Natural
  Language Processing of the AFNLP: Volume 1-Volume 1}, pages 513--521.
  Association for Computational Linguistics.

\bibitem[Kurata et~al., 2011]{kurata2011training}
Kurata, G., Itoh, N., and Nishimura, M. (2011).
\newblock Training of error-corrective model for {ASR} without using audio
  data.
\newblock In {\em Acoustics, Speech and Signal Processing (ICASSP), 2011 IEEE
  International Conference on}, pages 5576--5579. IEEE.

\bibitem[Ladhak et~al., 2016]{ladhak2016latticernn}
Ladhak, F., Gandhe, A., Dreyer, M., Mathias, L., Rastrow, A., and Hoffmeister,
  B. (2016).
\newblock Latticernn: Recurrent neural networks over lattices.
\newblock In {\em INTERSPEECH}, pages 695--699.

\bibitem[Lavie et~al., 1997]{lavie1997janus}
Lavie, A., Waibel, A., Levin, L., Finke, M., Gates, D., Gavalda, M.,
  Zeppenfeld, T., and Zhan, P. (1997).
\newblock Janus-iii: Speech-to-speech translation in multiple languages.
\newblock In {\em Acoustics, Speech, and Signal Processing, 1997. ICASSP-97.,
  1997 IEEE International Conference on}, volume~1, pages 99--102. IEEE.

\bibitem[Le and Mikolov, 2014]{le2014distributed}
Le, Q. and Mikolov, T. (2014).
\newblock Distributed representations of sentences and documents.
\newblock In {\em International Conference on Machine Learning}, pages
  1188--1196.

\bibitem[Levin et~al., 2013]{levin2013fixed}
Levin, K., Henry, K., Jansen, A., and Livescu, K. (2013).
\newblock Fixed-dimensional acoustic embeddings of variable-length segments in
  low-resource settings.
\newblock In {\em Automatic Speech Recognition and Understanding (ASRU), 2013
  IEEE Workshop on}, pages 410--415. IEEE.

\bibitem[Levy and Goldberg, 2014]{levy2014dependency}
Levy, O. and Goldberg, Y. (2014).
\newblock Dependency-based word embeddings.
\newblock In {\em Proceedings of the 52nd Annual Meeting of the Association for
  Computational Linguistics (Volume 2: Short Papers)}, volume~2, pages
  302--308.

\bibitem[Lilleberg et~al., 2015]{lilleberg2015support}
Lilleberg, J., Zhu, Y., and Zhang, Y. (2015).
\newblock Support vector machines and word2vec for text classification with
  semantic features.
\newblock In {\em Cognitive Informatics \& Cognitive Computing (ICCI* CC), 2015
  IEEE 14th International Conference on}, pages 136--140. IEEE.

\bibitem[Ling et~al., 2015]{ling2015two}
Ling, W., Dyer, C., Black, A.~W., and Trancoso, I. (2015).
\newblock Two/too simple adaptations of word2vec for syntax problems.
\newblock In {\em Proceedings of the 2015 Conference of the North American
  Chapter of the Association for Computational Linguistics: Human Language
  Technologies}, pages 1299--1304.

\bibitem[Liu et~al., 2014]{liu2014efficient}
Liu, X., Wang, Y., Chen, X., Gales, M.~J., and Woodland, P.~C. (2014).
\newblock Efficient lattice rescoring using recurrent neural network language
  models.
\newblock In {\em Acoustics, Speech and Signal Processing (ICASSP), 2014 IEEE
  International Conference on}, pages 4908--4912. IEEE.

\bibitem[Luong et~al., 2013]{luong2013better}
Luong, T., Socher, R., and Manning, C. (2013).
\newblock Better word representations with recursive neural networks for
  morphology.
\newblock In {\em Proceedings of the Seventeenth Conference on Computational
  Natural Language Learning}, pages 104--113.

\bibitem[Mangu et~al., 2000]{mangu2000finding}
Mangu, L., Brill, E., and Stolcke, A. (2000).
\newblock Finding consensus in speech recognition: word error minimization and
  other applications of confusion networks.
\newblock {\em Computer Speech \& Language}, 14(4):373--400.

\bibitem[Marin et~al., 2012]{marin2012using}
Marin, A., Kwiatkowski, T., Ostendorf, M., and Zettlemoyer, L. (2012).
\newblock Using syntactic and confusion network structure for out-of-vocabulary
  word detection.
\newblock In {\em Spoken Language Technology Workshop (SLT), 2012 IEEE}, pages
  159--164. IEEE.

\bibitem[Mathias and Byrne, 2006]{mathias2006statistical}
Mathias, L. and Byrne, W. (2006).
\newblock Statistical phrase-based speech translation.
\newblock In {\em Acoustics, Speech and Signal Processing, 2006. ICASSP 2006
  Proceedings. 2006 IEEE International Conference on}, volume~1, pages I--I.
  IEEE.

\bibitem[Matusov et~al., 2005]{matusov2005integration}
Matusov, E., Kanthak, S., and Ney, H. (2005).
\newblock On the integration of speech recognition and statistical machine
  translation.
\newblock In {\em Ninth European Conference on Speech Communication and
  Technology}.

\bibitem[Matusov et~al., 2006]{matusov2006computing}
Matusov, E., Ueffing, N., and Ney, H. (2006).
\newblock Computing consensus translation for multiple machine translation
  systems using enhanced hypothesis alignment.
\newblock In {\em 11th Conference of the European Chapter of the Association
  for Computational Linguistics}.

\bibitem[Mikolov et~al., 2013a]{mikolov2013efficient}
Mikolov, T., Chen, K., Corrado, G., and Dean, J. (2013a).
\newblock Efficient estimation of word representations in vector space.
\newblock {\em arXiv preprint arXiv:1301.3781}.

\bibitem[Mikolov et~al., 2010]{mikolov2010recurrent}
Mikolov, T., Karafi{\'a}t, M., Burget, L., {\v{C}}ernock{\`y}, J., and
  Khudanpur, S. (2010).
\newblock Recurrent neural network based language model.
\newblock In {\em Eleventh Annual Conference of the International Speech
  Communication Association}.

\bibitem[Mikolov et~al., 2013b]{mikolov2013exploiting}
Mikolov, T., Le, Q.~V., and Sutskever, I. (2013b).
\newblock Exploiting similarities among languages for machine translation.
\newblock {\em arXiv preprint arXiv:1309.4168}.

\bibitem[Mikolov et~al., 2013c]{mikolov2013distributed}
Mikolov, T., Sutskever, I., Chen, K., Corrado, G.~S., and Dean, J. (2013c).
\newblock Distributed representations of words and phrases and their
  compositionality.
\newblock In {\em Advances in neural information processing systems}, pages
  3111--3119.

\bibitem[Mnih and Kavukcuoglu, 2013]{mnih2013learning}
Mnih, A. and Kavukcuoglu, K. (2013).
\newblock Learning word embeddings efficiently with noise-contrastive
  estimation.
\newblock In {\em Advances in neural information processing systems}, pages
  2265--2273.

\bibitem[Niehues and Kolss, 2009]{niehues2009pos}
Niehues, J. and Kolss, M. (2009).
\newblock A pos-based model for long-range reorderings in smt.
\newblock In {\em Proceedings of the Fourth Workshop on Statistical Machine
  Translation}, pages 206--214. Association for Computational Linguistics.

\bibitem[Ogata and Goto, 2005]{ogata2005speech}
Ogata, J. and Goto, M. (2005).
\newblock Speech repair: Quick error correction just by using selection
  operation for speech input interfaces.
\newblock In {\em Ninth European Conference on Speech Communication and
  Technology}.

\bibitem[Onishi et~al., 2011]{onishi2011paraphrase}
Onishi, T., Utiyama, M., and Sumita, E. (2011).
\newblock Paraphrase lattice for statistical machine translation.
\newblock {\em IEICE TRANSACTIONS on Information and Systems},
  94(6):1299--1305.

\bibitem[Pennington et~al., 2014]{pennington2014glove}
Pennington, J., Socher, R., and Manning, C. (2014).
\newblock Glove: Global vectors for word representation.
\newblock In {\em Proceedings of the 2014 conference on empirical methods in
  natural language processing (EMNLP)}, pages 1532--1543.

\bibitem[Povey et~al., 2011]{povey2011kaldi}
Povey, D., Ghoshal, A., Boulianne, G., Burget, L., Glembek, O., Goel, N.,
  Hannemann, M., Motlicek, P., Qian, Y., Schwarz, P., Silovsky, J., Stemmer,
  G., and Vesely, K. (2011).
\newblock The kaldi speech recognition toolkit.
\newblock In {\em IEEE 2011 Workshop on Automatic Speech Recognition and
  Understanding}. IEEE Signal Processing Society.
\newblock IEEE Catalog No.: CFP11SRW-USB.

\bibitem[Qiu et~al., 2014]{qiu2014co}
Qiu, S., Cui, Q., Bian, J., Gao, B., and Liu, T.-Y. (2014).
\newblock Co-learning of word representations and morpheme representations.
\newblock In {\em Proceedings of COLING 2014, the 25th International Conference
  on Computational Linguistics: Technical Papers}, pages 141--150.

\bibitem[Quirk et~al., 2004]{quirk2004monolingual}
Quirk, C., Brockett, C., and Dolan, W. (2004).
\newblock Monolingual machine translation for paraphrase generation.
\newblock In {\em Proceedings of the 2004 Conference on Empirical Methods in
  Natural Language Processing}.

\bibitem[Rosti et~al., 2007a]{rosti2007combining}
Rosti, A.-V., Ayan, N.~F., Xiang, B., Matsoukas, S., Schwartz, R., and Dorr, B.
  (2007a).
\newblock Combining outputs from multiple machine translation systems.
\newblock In {\em Human Language Technologies 2007: The Conference of the North
  American Chapter of the Association for Computational Linguistics;
  Proceedings of the Main Conference}, pages 228--235.

\bibitem[Rosti et~al., 2007b]{rosti2007improved}
Rosti, A.-V., Matsoukas, S., and Schwartz, R. (2007b).
\newblock Improved word-level system combination for machine translation.
\newblock In {\em Proceedings of the 45th Annual Meeting of the Association of
  Computational Linguistics}, pages 312--319.

\bibitem[Sagae et~al., 2012]{sagae2012hallucinated}
Sagae, K., Lehr, M., Prud'hommeaux, E., Xu, P., Glenn, N., Karakos, D.,
  Khudanpur, S., Roark, B., Saraclar, M., Shafran, I., et~al. (2012).
\newblock Hallucinated n-best lists for discriminative language modeling.
\newblock In {\em Acoustics, Speech and Signal Processing (ICASSP), 2012 IEEE
  International Conference on}, pages 5001--5004. IEEE.

\bibitem[Schnabel et~al., 2015]{schnabel2015evaluation}
Schnabel, T., Labutov, I., Mimno, D., and Joachims, T. (2015).
\newblock Evaluation methods for unsupervised word embeddings.
\newblock In {\em Proceedings of the 2015 Conference on Empirical Methods in
  Natural Language Processing}, pages 298--307.

\bibitem[Schroeder et~al., 2009]{schroeder2009word}
Schroeder, J., Cohn, T., and Koehn, P. (2009).
\newblock Word lattices for multi-source translation.
\newblock In {\em Proceedings of the 12th Conference of the European Chapter of
  the Association for Computational Linguistics}, pages 719--727. Association
  for Computational Linguistics.

\bibitem[Schultz et~al., 2004]{schultz2004using}
Schultz, T., Jou, S.-C., Vogel, S., and Saleem, S. (2004).
\newblock Using word latice information for a tighter coupling in speech
  translation systems.
\newblock In {\em Eighth International Conference on Spoken Language
  Processing}.

\bibitem[Seigel and Woodland, 2011]{seigel2011combining}
Seigel, M.~S. and Woodland, P.~C. (2011).
\newblock Combining information sources for confidence estimation with crf
  models.
\newblock In {\em Twelfth Annual Conference of the International Speech
  Communication Association}.

\bibitem[Shivakumar et~al., 2018]{shivakumar2018learning}
Shivakumar, P.~G., Li, H., Knight, K., and Georgiou, P. (2018).
\newblock Learning from past mistakes: Improving automatic speech recognition
  output via noisy-clean phrase context modeling.
\newblock {\em arXiv preprint arXiv:1802.02607}.

\bibitem[Soricut and Och, 2015]{soricut2015unsupervised}
Soricut, R. and Och, F. (2015).
\newblock Unsupervised morphology induction using word embeddings.
\newblock In {\em Proceedings of the 2015 Conference of the North American
  Chapter of the Association for Computational Linguistics: Human Language
  Technologies}, pages 1627--1637.

\bibitem[Sperber et~al., 2017]{sperber2017neural}
Sperber, M., Neubig, G., Niehues, J., and Waibel, A. (2017).
\newblock Neural lattice-to-sequence models for uncertain inputs.
\newblock {\em arXiv preprint arXiv:1704.00559}.

\bibitem[Su et~al., 2017]{su2017lattice}
Su, J., Tan, Z., Xiong, D., Ji, R., Shi, X., and Liu, Y. (2017).
\newblock Lattice-based recurrent neural network encoders for neural machine
  translation.
\newblock In {\em AAAI}, pages 3302--3308.

\bibitem[Sundermeyer et~al., 2014]{sundermeyer2014lattice}
Sundermeyer, M., T{\"u}ske, Z., Schl{\"u}ter, R., and Ney, H. (2014).
\newblock Lattice decoding and rescoring with long-span neural network language
  models.
\newblock In {\em Fifteenth Annual Conference of the International Speech
  Communication Association}.

\bibitem[Tai et~al., 2015]{tai2015improved}
Tai, K.~S., Socher, R., and Manning, C.~D. (2015).
\newblock Improved semantic representations from tree-structured long
  short-term memory networks.
\newblock {\em arXiv preprint arXiv:1503.00075}.

\bibitem[Tan et~al., 2010]{tan2010automatic}
Tan, Q.~F., Audhkhasi, K., Georgiou, P.~G., Ettelaie, E., and Narayanan, S.~S.
  (2010).
\newblock Automatic speech recognition system channel modeling.
\newblock In {\em Eleventh Annual Conference of the International Speech
  Communication Association}.

\bibitem[Tan et~al., 2018]{tan2018lattice}
Tan, Z., Su, J., Wang, B., Chen, Y., and Shi, X. (2018).
\newblock Lattice-to-sequence attentional neural machine translation models.
\newblock {\em Neurocomputing}, 284:138--147.

\bibitem[Tur et~al., 2002]{tur2002improving}
Tur, G., Wright, J., Gorin, A., Riccardi, G., and Hakkani-T{\"u}r, D. (2002).
\newblock Improving spoken language understanding using word confusion
  networks.
\newblock In {\em Seventh International Conference on Spoken Language
  Processing}.

\bibitem[Wahlster, 2013]{wahlster2013verbmobil}
Wahlster, W. (2013).
\newblock {\em Verbmobil: foundations of speech-to-speech translation}.
\newblock Springer Science \& Business Media.

\bibitem[Weide, 1998]{weide1998cmu}
Weide, R. (1998).
\newblock The cmu pronunciation dictionary, release 0.6.

\bibitem[Wuebker and Ney, 2012]{wuebker2012phrase}
Wuebker, J. and Ney, H. (2012).
\newblock Phrase model training for statistical machine translation with word
  lattices of preprocessing alternatives.
\newblock In {\em Proceedings of the Seventh Workshop on Statistical Machine
  Translation}, pages 450--459. Association for Computational Linguistics.

\bibitem[Xing et~al., 2014]{xing2014document}
Xing, C., Wang, D., Zhang, X., and Liu, C. (2014).
\newblock Document classification with distributions of word vectors.
\newblock In {\em Signal and Information Processing Association Annual Summit
  and Conference (APSIPA), 2014 Asia-Pacific}, pages 1--5. IEEE.

\bibitem[Xiong et~al., 2016]{xiong2016achieving}
Xiong, W., Droppo, J., Huang, X., Seide, F., Seltzer, M., Stolcke, A., Yu, D.,
  and Zweig, G. (2016).
\newblock Achieving human parity in conversational speech recognition.
\newblock {\em arXiv preprint arXiv:1610.05256}.

\bibitem[Xu et~al., 2011]{xu2011minimum}
Xu, H., Povey, D., Mangu, L., and Zhu, J. (2011).
\newblock Minimum bayes risk decoding and system combination based on a
  recursion for edit distance.
\newblock {\em Computer Speech \& Language}, 25(4):802--828.

\bibitem[Xu et~al., 2012]{xu2012phrasal}
Xu, P., Roark, B., and Khudanpur, S. (2012).
\newblock Phrasal cohort based unsupervised discriminative language modeling.
\newblock In {\em Thirteenth Annual Conference of the International Speech
  Communication Association}.

\bibitem[Xue and Zhao, 2005]{xue2005improved}
Xue, J. and Zhao, Y. (2005).
\newblock Improved confusion network algorithm and shortest path search from
  word lattice.
\newblock In {\em Acoustics, Speech, and Signal Processing, 2005.
  Proceedings.(ICASSP'05). IEEE International Conference on}, volume~1, pages
  I--853. IEEE.

\bibitem[Yin and Sch{\"{u}}tze, 2016]{YinS16}
Yin, W. and Sch{\"{u}}tze, H. (2016).
\newblock Learning word meta-embeddings.
\newblock In {\em Proceedings of the 54th Annual Meeting of the Association for
  Computational Linguistics, {ACL} 2016, August 7-12, 2016, Berlin, Germany,
  Volume 1: Long Papers}.

\end{thebibliography}

\newpage

\setcounter{table}{0}
\renewcommand{\thetable}{A\arabic{table}}
\section*{Appendix}
\begin{table*}[ht]
\begin{center}
\resizebox{\linewidth}{!}{%
\begin{tabular}{|l|c|c|c|c|c|c|c|c|}
\hline
\multirow{3}{*}{Model} & \multicolumn{8}{|c|}{Analogy Tasks} \\
\cline{2-9}
& \multicolumn{3}{|c|}{Semantic\&Syntactic Analogy} & \multirow{2}{*}{Acoustic Analogy} & \multicolumn{3}{|c|}{Semantic\&Syntactic-Acoustic Analogy} & \multirow{2}{*}{Average Accuracy} \\
\cline{2-4}
\cline{6-8}
& Semantic & Syntactic & Semantic\&Syntactic & & Semantic-Acoustic & Syntactic-Acoustic & Semantic\&Syntactic-Acoustic & \\
\hline
Google W2V & 28.98\% (35.75\%) & 70.79\% (78.74\%) & 61.42\% (69.1\%) & 0.9\% (1.42\%) & 6.54\% (14.38\%) & 17.9\% (27.46\%) & 16.99\% (26.42\%) & 26.44\% (32.31\%) \\
In-domain W2V & 42.39\% (51.57\%) & 33.14\% (43.14\%) & 35.15\% (44.98\%) & 0.3\% (0.6\%) & 5.17\% (10.69\%) & 8.13\% (11.93\%) & 7.86\% (11.82\%) & 14.44\% (19.13\%) \\
C2V-1 & 38.33\% (46.7\%) & 33.1\% (42.36\%) & 34.27\% (43.33\%) & 0.7\% (1.16\%) & 11.76\% (14.38\%) & 11.23\% (15.11\%) & 11.27\% (15.05\%) & 15.41\% (19.85\%) \\
C2V-a & 0.51\% (0.78\%) & 18.59\% (28.17\%) & 14.54\% (22.03\%) & 41.93\% (52.58\%) & 0.98\% (2.29\%) & 9.62\% (15.67\%) & 8.94\% (14.61\%) & 21.8\% (29.74\%) \\
C2V-c & 16.15\% (23.7\%) & 26.14\% (39.74\%) & 23.9\% (36.15\%) & 48.58\% (60.57\%) & 3.27\% (6.86\%) & 12.13\% (21.61\%) & 11.42\% (20.44\%) & 27.97\% (39.05\%) \\
C2V-* & 2.07\% (2.58\%) & 28.91\% (38.6\%) & 22.89\% (30.53\%) & 40.78\% (53.55\%) & 1.96\% (2.94\%) & 20.99\% (31.63\%) & 19.48\% (29.35\%) & 27.72\% (37.81\%) \\
\hline
\end{tabular}%
}
\end{center}
\captionsetup{justification=centering}
\caption{\textbf{Analogy Task Results with Semantic \& Syntactic splits: Different proposed models}\\
\small{\textbf{C2V-1: Top-Confusion, C2V-a: Intra-Confusion, C2V-c: Inter-Confusion, C2V-*: Hybrid Intra-Inter}\\
All the models are of 256 dimensions except Google W2V (300 dimensions).\\
Numbers inside parenthesis indicate top-2 evaluation accuracy;\\
Numbers outside parenthesis represent top-1 evaluation accuracy.\\
Google Word2Vec, Word2Vec Groundtruth (trained on in-domain) and Baseline Word2Vec (trained on ASR transcriptions) perform better with the Semantic\&Syntactic tasks, but fares poorly with Acoustic analogy task.\\
Intra-Confusion performs well on Acoustic analogy task while compromising on Semantic\&Syntactic task.\\
Inter-Confusion performs well on both the Acoustic analogy and Semantic\&Syntactic tasks.\\
Hybrid Intra-Inter training performs fairly well on all the three analogy tasks (Acoustic, Semantic\&Syntactic and Semantic\&Syntactic-Acoustic).}}\label{appendix:tab:results}
\end{table*}

\begin{table*}[hb]
\begin{center}
\setlength{\tabcolsep}{12pt}
\begin{tabular}{|l|c|c|}
\hline
\multirow{2}{*}{Model} & \multicolumn{2}{|c|}{Similarity Tasks} \\
\cline{2-3}
& Word Similarity & Acoustic Similarity \\
\hline
Google W2V & 0.6893 (7.9e-48) & -0.3489 (2.2e-28) \\
In-domain W2V & 0.5794 (4.2e-29) & -0.2444 (1e-10) \\
C2V-1 & 0.4992 (3.3e-22) & 0.1944 (1.7e-9) \\
C2V-a & 0.105 (0.056) & 0.8138 (5.1e-224) \\
C2V-c & 0.2937 (5.4e-8) & 0.8055 (5.1e-216) \\
C2V-* & 0.0963 (0.08) & 0.7858 (1.5e-198) \\
\hline
\end{tabular}%
\end{center}
\captionsetup{justification=centering}
\caption{\textbf{Similarity Task Results: Different proposed models}\\
\small{\textbf{C2V-1: Top-Confusion, C2V-a: Intra-Confusion, C2V-c: Inter-Confusion, C2V-*: Hybrid Intra-Inter}\\Similarity in terms of Spearman's correlation.\\
All the models are of 256 dimensions except Google W2V (300 dimensions).\\
Numbers inside parenthesis indicate correlation $p-value$ for similarity tasks\\
Google Word2Vec, Baseline Word2Vec and Word2Vec Groundtruth, all show high correlations with word similarity, while showing poor correlations on acoustic similarity. Google Word2Vec and Word2Vec Groundtruth models trained on clean data exhibit negative acoustic similarity correlation. Baseline Word2Vec trained on noisy ASR shows a small positive acoustic similarity correlation.\\
Intra-Confusion, Inter-Confusion and Hybrid Intra-Inter training show higher correlations on Acoustic similarity.}}
\label{appendix:tab:results:similarity}
\end{table*}

\begin{table*}[ht]
\begin{center}
\resizebox{\linewidth}{!}{%
\begin{tabular}{|l|c|c|c|c|c|c|c|c|}
\hline
\multirow{3}{*}{Model} & \multicolumn{8}{|c|}{Analogy Tasks} \\
\cline{2-9}
& \multicolumn{3}{|c|}{Semantic\&Syntactic Analogy} & \multirow{2}{*}{Acoustic Analogy} & \multicolumn{3}{|c|}{Semantic\&Syntactic-Acoustic Analogy} & \multirow{2}{*}{Average Accuracy} \\
\cline{2-4}
\cline{6-8}
& Semantic & Syntactic & Semantic\&Syntactic & & Semantic-Acoustic & Syntactic-Acoustic & Semantic\&Syntactic-Acoustic & \\
\hline
Google W2V & 28.98\% (35.75\%) & 70.79\% (78.74\%) & 61.42\% (69.1\%) & 0.9\% (1.42\%) & 6.54\% (14.38\%) & 17.9\% (27.46\%) & 16.99\% (26.42\%) & 26.44\% (32.31\%) \\
In-domain W2V & 32.72\% (39.99\%) & 66.53\% (75.97\%) & 59.17\% (68.14\%) & 0.6\% (0.96\%) & 10.52\% (17.46\%) & 10.5\% (17.69\%) & 8.15\% (13.5\%) & 22.64\% (27.53\%) \\
C2V-1 & 34.92\% (41.96\%) & 68.7\% (78.82\%) & 61.13\% (70.56\%) & 0.9\% (1.46\%) & 14.38\% (19.28\%) & 16.85\% (24.25\%) & 16.66\% (23.86\%) & 26.23\% (31.96\%) \\
C2V-a & 11.5\% (15.53\%) & 67.56\% (77.96\%) & 54.99\% (63.97\%) & 9.04\% (16.92\%) & 7.84\% (10.46\%) & 36.92\% (46.17\%) & 34.61\% (43.34\%) & 32.88\% (41.41\%) \\
C2V-c & 25.77\% (33.12\%) & 60.1\% (74.79\%) & 52.4\% (65.45\%) & 16.54\% (27.33\%) & 10.78\% (14.05\%) & 28.9\% (40.38\%) & 27.46\% (38.29\%) & 32.13\% (43.69\%) \\
C2V-* & 15.64\% (21.94\%) & 66.73\% (77.68\%) & 55.28\% (65.19\%) & 10.49\% (20.35\%) & 6.86\% (11.11\%) & 35.4\% (44.85\%) & 33.13\% (42.18\%) & 36.27\% (42.57\%) \\
\hline
\end{tabular}%
}
\end{center}
\captionsetup{justification=centering}
\caption{\textbf{Analogy Task Results with Semantic \& Syntactic splits: Model pre-training/initialization}\\
\small{\textbf{C2V-1: Top-Confusion, C2V-a: Intra-Confusion, C2V-c: Inter-Confusion, C2V-*: Hybrid Intra-Inter}\\
All the models are of 300 dimensions. Numbers inside parenthesis indicate top-2 evaluation accuracy;\\
Numbers outside parenthesis represent top-1 evaluation accuracy.\\
Pre-training is helpful in all the cases. Pre-training boosts the Semantic\&Syntactic Analogy accuracy for all.\\
For Intra-Confusion, Inter-Confusion and Hybrid Intra-Inter models, pre-training boosts the Semantic\&Syntactic-Acoustic Analogy accuracies. A small dip in Acoustic Analogy accuracies is observed. However, overall average accuracy is improved.}}\label{appendix:tab:pretraining_results}
\end{table*}

\begin{table*}[ht]
\begin{center}
\setlength{\tabcolsep}{12pt}
\begin{tabular}{|l|c|c|}
\hline
\multirow{2}{*}{Model} & \multicolumn{2}{|c|}{Similarity Tasks} \\
\cline{2-3}
& Word Similarity & Acoustic Similarity \\
\hline
Google W2V & 0.6893 (7.9e-48) & -0.3489 (2.2e-28) \\
In-domain W2V & 0.4417 (3.5e-16) & -0.4377 (3.6e-33)  \\
C2V-1 & 0.6036 (3.8e-34) & -0.4327 (2.5e-44) \\
C2V-a & 0.5228 (1.4e-24) & 0.62 (2.95e-101) \\
C2V-c & 0.5798 (4.9e-31) & 0.5825 (9.1e-87) \\
C2V-* & 0.5341 (9.8e-26) & 0.6237 (8.8e-103) \\
\hline
\end{tabular}%
\end{center}
\captionsetup{justification=centering}
\caption{\textbf{Similarity Task Results: Model pre-training/initialization}\\
\small{\textbf{C2V-1: Top-Confusion, C2V-a: Intra-Confusion, C2V-c: Inter-Confusion, C2V-*: Hybrid Intra-Inter}\\Similarity in terms of Spearman's correlation.\\
All the models are of 300 dimensions.\\
 Numbers inside parenthesis indicate correlation $p-value$ for similarity tasks.\\
Pre-training boosts the Word Similarity correlation for all the models. The correlation is improved considerably in the case of Intra-Confusion, Inter-Confusion and Hybrid Intra-Inter models while maintaining good correlation on acoustic similarity.}}
\label{appendix:tab:pretraining_results:similarity}
\end{table*}

\begin{table*}[ht]
\begin{center}
\resizebox{\linewidth}{!}{%
\begin{tabular}{|l|c|c|c|c|c|c|c|c|c|}
\hline
\multirow{3}{*}{Model} & \multirow{3}{*}{Fine-tuning} & \multicolumn{8}{|c|}{Analogy Tasks} \\
\cline{3-10}
& & \multicolumn{3}{|c|}{Semantic\&Syntactic Analogy} & \multirow{2}{*}{Acoustic Analogy} & \multicolumn{3}{|c|}{Semantic\&Syntactic-Acoustic Analogy} & \multirow{2}{*}{Average Accuracy} \\
\cline{3-5}\cline{7-9}
& Scheme & Semantic & Syntactic & Semantic\&Syntactic & & Semantic-Acoustic & Syntactic-Acoustic & Semantic\&Syntactic-Acoustic & \\
\hline
Google W2V & - & 28.98\% (35.75\%) & 70.79\% (78.74\%) & 61.42\% (69.1\%) & 0.9\% (1.42\%) & 6.54\% (14.38\%) & 17.9\% (27.46\%) & 16.99\% (26.42\%) & 26.44\% (32.31\%) \\
In-domain W2V (556 dim.) & - & 39.11\% (48.03\%) & 70.41\% (79.54\%) & 63.6\% (72.69\%) & 0.81\% (1.0\%) & 12.07\% (18.62\%) & 14.79\% (24.91\%) & 14.54\% (24.33\%) & 26.32\% (32.67\%) \\
\hline
\multicolumn{10}{c}{Model Concatenation} \\
\hline
C2V-1 (F) + C2V-a (F) & - & 6.22\% (9.5\%) & 71.03\% (83.65\%) & 56.51\% (67.03\%) & 13.59\% (25.43\%) & 6.54\% (11.76\%) & 33.91\% (42.82\%) & 31.74\% (40.36\%) & 33.95\% (44.27\%) \\
C2V-1 (F) + C2V-c (F) & - & 36.53\% (47.01\%) & 57.94\% (77.72\%) & 53.14\% (70.84\%) & 20.99\% (35.25\%) & 10.46\% (16.01\%) & 26.31\% (36.83\%) & 25.05\% (35.18\%) & 33.06\% (47.09\%) \\
C2V-1 (F) + C2V-* (F) & - & 11.85\% (17.32\%) & 71.85\% (82.74\%) & 58.4\% (68.08\%) & 6.35\% (11.39\%) & 7.84\% (12.18\%) & 34.38\% (43.78\%) & 32.28\% (41.3\%) & 32.34\% (40.26\%) \\
\hline
\multicolumn{10}{c}{Fixed Contextual Subspace Joint Optimization} \\
\hline
C2V-1 (F) + C2V-a (L) & inter & 22.96\% (32.42\%) & 66.19\% (82.98\%) & 56.5\% (71.65\%) & 12.73\% (20.54\%) & 13.4\% (18.3\%) & 26.22\% (35.09\%) & 25.21\% (33.76\%) &  31.48\% (41.98\%) \\
C2V-1 (F) + C2V-a (L) & intra & 6.69\% (11.58\%) & 69.79\% (83.48\%) & 55.65\% (67.37\%) & 17.03\% (28.64\%) & 8.17\% (13.73\%) & 31.85\% (47.64\%) & 29.97\% (39.09\%) & 34.22\% (45.03\%) \\
C2V-1 (F) + C2V-a (L) & hybrid & 11.69\% (19.79\%) & 69.31\% (84.53\%) & 56.39\% (70.02\%) & 14.86\% (25.84\%) & 9.8\% (16.67\%) & 30.02\% (38.94\%) & 28.42\% (37.18\%) & 33.22\% (44.35\%) \\
C2V-1 (F) + C2V-c (L) & inter & 39.19\% (50.57\%) & 58.35\% (78.21\%) & 54.05\% (72.01\%) & 23.33\% (35.25\%) & 12.42\% (18.3\%) & 24.45\% (34.89\%) & 23.5\% (33.58\%) & 33.63\% (46.95\%) \\
C2V-1 (F) + C2V-c (L) & intra & 22.76\% (32.85\%) & 62.07\% (80.34\%) & 53.26\% (69.7\%) & 24.76\% (39.32\%) & 7.52\% (11.11\%) & 29.97\% (41.47\%) & 28.19\% (39.07\%) & 35.40\% (49.36\%) \\
C2V-1 (F) + C2V-c (L) & hybrid & 30.54\% (43.21\%) & 61.56\% (80.81\%) & 54.61\% (72.38\%) & 23.6\% (37.75\%) & 8.5\% (14.71\%) & 28.25\% (39.95\%) & 26.68\% (37.95\%) & 34.96\% (49.36\%) \\
C2V-1 (F) + C2V-* (L) & inter & 27.02\% (35.9\%) & 67.52\% (81.6\%) & 58.45\% (71.36\%) & 5.04\% (8.55\%) & 11.76\% (16.67\%) & 26.28\% (34.64\%) & 25.13\% (33.21\%) & 29.54\% (37.71\%) \\
C2V-1 (F) + C2V-* (L) & intra & 10.48\% (15.84\%) & 70.44\% (81.57\%) & 57.00\% (66.85\%) & 7.21\% (13.33\%) & 6.21\% (12.09\%) & 34.07\% (42.52\%) & 31.87\% (40.1\%) & 32.03\% (40.09) \\
C2V-1 (F) + C2V-* (L) & hybrid & 15.41\% (23.31\%) & 70.56\% (82.61\%) & 58.2\% (68.32\%) & 6.39\% (11.61\%) & 8.17\% (12.09\%) & 32.36\% (40.43\%) & 30.44\% (38.19\%) & 31.68\% (39.37\%) \\
\hline
\multicolumn{10}{c}{Unrestricted Joint Optimization} \\
\hline
C2V-1 (L) + C2V-a (L) & inter & 8.6\% (14.74\%) & 57.96\% (75.8\%) & 46.9\% (62.12\%) & 30.73\% (46.42\%) & 5.88\% (12.75\%) & 26.79\% (38.44\%) & 25.13\% (36.4\%) & 34.25\% (48.31\%) \\
C2V-1 (L) + C2V-a (L) & intra & 4.97\% (7.9\%) & 69.27\% (81.30\%) & 54.86\% (64.85\%) & 23.86\% (40.55\%) & 7.84\% (11.44\%) & 34.92\% (45.02\%) & 32.77\% (42.38\%) & 37.16\% (49.26\%) \\
C2V-1 (L) + C2V-a (L) & hybrid & 1.1\% (1.64\%) & 26.54\% (40.32\%) & 20.83\% (31.65\%) & 49.25\% (61.91\%) & 2.29\% (3.92\%) & 15.05\% (25.24\%) & 14.04\% (23.55\%) & 28.12\% (39.04\%) \\
C2V-1 (L) + C2V-c (L) & inter & 33.01\% (43.72\%) & 50.81\% (71.13\%) & 46.82\% (64.98\%) & 37.15\% (52.99\%) & 9.48\% (16.01\%) & 23.16\% (36.41\%) & 22.07\% (34.79\%) & 35.35\% (50.92\%) \\
C2V-1 (L) + C2V-c (L) & intra & 21.9\% (30.43\%) & 58.99\% (76.12\%) & 50.68\% (65.88\%) & 33.05\% (49.4\%) & 7.52\% (10.46\%) & 31.23\% (44.12\%) & 29.35\% (41.51\%) & 37.69\% (52.26\%) \\
C2V-1 (L) + C2V-c (L) & hybrid &  10.48\% (15.72\%) & 30.0\% (44.25\%) & 25.63\% (37.86\%) & 52.73\% (67.21\%) & 3.27\% (4.9\%) & 16.09\% (27.77\%) & 15.08\% (25.96\%) & 31.15\% (43.68\%) \\
C2V-1 (L) + C2V-* (L) & inter & 19.24\% (26.59\%) & 61.57\% (76.8\%) & 52.08\% (65.54\%) & 17.85\% (27.97\%) & 7.52\% (12.75\%) & 28.81\% (38.94\%) & 27.12\% (36.87\%) & 32.35\% (43.46\%) \\
C2V-1 (L) + C2V-* (L) & intra & 10.09\% (13.77\%) & 68.76\% (79.06\%) & 55.61\% (64.42\%) & 10.34\% (20.05\%) & 5.88\% (9.48\%) & 36.13\% (45.41\%) & 33.73\% (42.56\%) & 33.23\% (42.34\%) \\
C2V-1 (L) + C2V-* (L) & hybrid & 12.98\% (17.91\%) & 68.26\% (79.62\%) & 55.87\% (65.79\%) & 11.73\% (22.63\%) & 5.88\% (10.46\%) & 35.28\% (43.92\%) & 32.95\% (41.3\%) & 33.52\% (43.24\%) \\
\hline
\end{tabular}%
}
\end{center}
\captionsetup{justification=centering}
\caption{\textbf{Analogy Task Results: Model concatenation and joint optimization}\\
\small{\textbf{C2V-1: Top-Confusion, C2V-a: Intra-Confusion, C2V-c: Inter-Confusion, C2V-*: Hybrid Intra-Inter}\\Numbers inside parenthesis indicate top-2 evaluation accuracy;\\
All the models are of 556 dimensions.\\
 Numbers outside parenthesis represent top-1 evaluation accuracy.\\
Acronyms: (F):Fixed embedding, (L):Learn embedding during joint training\\
Model Concatenation provides gains in Acoustic-Analogy Task and thereby resulting in gains in average accuracy compared to results in Table~\ref{appendix:tab:pretraining_results} for Intra-Confusion and Inter-Confusion models.\\
Fixed Contextual Subspace and Unrestricted Joint Optimizations further improves results over model concatenation. Best results in terms of average accuracy is obtained with unrestricted joint optimizations, an absolute improvement of 10\%.\\
Confusion2Vec models surpass Word2Vec even for Semantic\&Syntactic analogy task (top-2 evaluation accuracy).}}\label{appendix:tab:joint_opt_results}
\end{table*}

\begin{table*}[ht]
\begin{center}
\resizebox{\linewidth}{!}{%
\setlength{\tabcolsep}{12pt}
\begin{tabular}{|l|c|c|c|}
\hline
\multirow{2}{*}{Model} & Fine-tuning & \multicolumn{2}{|c|}{Similarity Tasks} \\
\cline{3-4}
& Scheme & Word Similarity & Acoustic Similarity \\
\hline
Google W2V & - & 0.6893 (7.9e-48) & -0.3489 (2.2e-28) \\
In-domain W2V (556 dim.) & - & 0.6333 (4.9e-e36) & -0.4717 (5.7e-39) \\
\hline
\multicolumn{4}{c}{Model Concatenation} \\
\hline
C2V-1 (F) + C2V-a (F) & - & 0.5102 (2.9e-23) & 0.7231 (2.2e-153) \\
C2V-1 (F) + C2V-c (F) & - & 0.5609 (9.8e-29) & 0.6345 (2.3e-107) \\
C2V-1 (F) + C2V-* (F) & - & 0.4142 (4.1e-15) & 0.5285 (5.6e-69) \\
\hline
\multicolumn{4}{c}{Fixed Contextual Subspace Joint Optimization} \\
\hline
C2V-1 (F) + C2V-a (L) & inter & 0.5676 (1.6e-29) & 0.4437 (9.1e-47) \\
C2V-1 (F) + C2V-a (L) & intra & 0.5211 (2.3e-24) & 0.6967 (6.5e-138) \\
C2V-1 (F) + C2V-a (L) & hybrid & 0.5384 (3.4e-26) & 0.6287 (6.7e-105) \\
C2V-1 (F) + C2V-c (L) & inter & 0.5266 (6.1e-25) & 0.5818 (1.6e-86) \\
C2V-1 (F) + C2V-c (L) & intra & 0.5156 (8.3e-24) & 0.7021 (6.3e-141) \\
C2V-1 (F) + C2V-c (L) & hybrid & 0.5220 (1.8e-24) & 0.6674 (1.4e-122) \\
C2V-1 (F) + C2V-* (L) & inter & 0.5587 (1.7e-28) & 0.302 (2.5e-21) \\
C2V-1 (F) + C2V-* (L) & intra & 0.4996 (3.1e-22) & 0.5691 (4.7e-82) \\
C2V-1 (F) + C2V-* (L) & hybrid & 0.5254 (8.2e-25) & 0.4945 (2.6e-59) \\
\hline
\multicolumn{4}{c}{Unrestricted Joint Optimization} \\
\hline
C2V-1 (L) + C2V-a (L) & inter & 0.5513 (1.3e-27) & 0.7926 (2.4e-204) \\
C2V-1 (L) + C2V-a (L) & intra & 0.5033 (1.4e-22) & 0.7949 (2e-206) \\
C2V-1 (L) + C2V-a (L) & hybrid & 0.1067 (0.0528) & 0.8309 (8.5e-242) \\
C2V-1 (L) + C2V-c (L) & inter & 0.5763 (1.3e-30) & 0.7725 (8.2e-188) \\
C2V-1 (L) + C2V-c (L) & intra & 0.5379 (3.8e-26) & 0.7717 (3.5e-187) \\
C2V-1 (L) + C2V-c (L) & hybrid & 0.2295 (2.6e-5) & 0.8294 (3.6e-240) \\
C2V-1 (L) + C2V-* (L) & inter & 0.5338 (1e-25) & 0.6953 (3.7e-137) \\
C2V-1 (L) + C2V-* (L) & intra & 0.4920 (1.6e-21) & 0.6942 (1.5e-136) \\
C2V-1 (L) + C2V-* (L) & hybrid & 0.4967 (5.8e-22) & 0.6986 (5.9e-139) \\
\hline
\end{tabular}%
}
\end{center}
\captionsetup{justification=centering}
\caption{\textbf{Similarity Task Results: Model concatenation and joint optimization}\\
\small{\textbf{C2V-1: Top-Confusion, C2V-a: Intra-Confusion, C2V-c: Inter-Confusion, C2V-*: Hybrid Intra-Inter}\\Similarity in terms of Spearman's correlation.\\
All the models are of 556 dimensions.\\
Numbers inside parenthesis indicate correlation $p-value$ for similarity tasks.\\
Good correlations are observed for both the word similarity and acoustic similarity with model concatenation with and without joint optimization. All the correlations are found to be statistically significant.}}
\label{appendix:tab:joint_opt_results:similarity}
\end{table*}

\end{document}